\pdfoutput=1

\documentclass[11pt]{article}

\usepackage[]{acl}

\usepackage{times}
\usepackage{latexsym}

\usepackage[T1]{fontenc}

\usepackage[utf8]{inputenc}

\usepackage{microtype}

\usepackage{inconsolata}
\usepackage{dsfont}
\usepackage{amsmath}

\usepackage{graphicx}
\usepackage{amssymb}

\usepackage{multirow}
\usepackage{booktabs}
\usepackage{xcolor}
\usepackage{subcaption}
\definecolor{lightblue}{RGB}{0,127,255}
\definecolor{yellow-history}{RGB}{234,216,91}
\definecolor{green-diagnosis}{RGB}{86, 181, 170}
\definecolor{blue-retrieval}{RGB}{116, 167, 226}

\usepackage{enumitem}

\usepackage[most]{tcolorbox}

\newtcolorbox{prompt}[2][]{
    colback=black!3,
    colframe=black,
    boxrule=0.5pt,
    left=8pt,
    right=8pt,
    top=8pt,
    bottom=8pt,
    arc=2pt,
    breakable,
    enhanced,
    before skip=10pt,
    after skip=10pt,
    title={#2}, 
    #1 
}

\newtcolorbox{example}[2][]{
    colback=black!3,
    colframe=black!20,
    boxrule=0.5pt,
    left=8pt,
    right=8pt,
    top=8pt,
    bottom=8pt,
    arc=2pt,
    breakable,
    enhanced,
    before skip=10pt,
    after skip=10pt,
    coltitle=black,
    title={#2}, 
    #1 
}

\title{MEDDxAgent: A Unified Modular Agent Framework for\\ Explainable Automatic Differential Diagnosis}

\author{Daniel Rose\textsuperscript{1}, Chia-Chien Hung\textsuperscript{2}, Marco Lepri\textsuperscript{2}, \\
\textbf{Israa Alqassem\textsuperscript{2}}, \textbf{Kiril Gashteovski\textsuperscript{2, 3}} and \textbf{Carolin Lawrence\textsuperscript{2}} \\
\textsuperscript{1}University of California, Santa Barbara \\
\textsuperscript{2}NEC Laboratories Europe, Heidelberg, Germany\\
\textsuperscript{3}CAIR, Ss. Cyril and Methodius University of Skopje, North Macedonia \\
  \texttt{danielrose@ucsb.edu}\\
  \texttt{\{Chia-Chien.Hung, Marco.Lepri,} \\
  \texttt{Israa.Alqassem, Kiril.Gashteovski, Carolin.Lawrence\}@neclab.eu}\\
  }
  
\begin{document}
\maketitle
\begin{abstract}

Differential Diagnosis (DDx) is a fundamental yet complex aspect of clinical decision-making, in which physicians iteratively refine a ranked list of possible diseases based on symptoms, antecedents, and medical knowledge. While recent advances in large language models (LLMs) have shown promise in supporting DDx, existing approaches face key limitations, including single-dataset evaluations, isolated optimization of components, unrealistic assumptions about complete patient profiles, and single-attempt diagnosis. We introduce a \textbf{M}odular \textbf{E}xplainable \textbf{DDx} \textbf{Agent} (\textbf{MEDDxAgent}) framework designed for interactive DDx, where diagnostic reasoning evolves through \textit{iterative learning}, rather than assuming a complete patient profile is accessible. MEDDxAgent integrates three modular components: (1) an orchestrator (DDxDriver), (2) a history taking simulator, and (3) two specialized agents for knowledge retrieval and diagnosis strategy. To ensure robust evaluation, we introduce a comprehensive DDx benchmark covering respiratory, skin, and rare diseases.  We analyze single-turn diagnostic approaches and demonstrate the importance of iterative refinement when patient profiles are not available at the outset. Our broad evaluation demonstrates that MEDDxAgent achieves over 10\% accuracy improvements in interactive DDx across both large and small LLMs, while offering critical explainability into its diagnostic reasoning process.  

\end{abstract}


\section{Introduction}
Differential Diagnosis (DDx) is a crucial step in medical decision-making, where doctors systematically narrow down the most likely diagnosis from a range of possible diseases \citep{rhoads2017formulating}. In real-world clinical practice, DDx is essential because it accounts for uncertainty in the diagnosis \citep{henderson2012patient}. It's also incredibly challenging given the large number of potential diseases, rapidly evolving medical knowledge, and the fact that symptoms and antecedents can point to multiple diseases \citep{winter2024ddxgym}.
Expert clinicians rely on pattern recognition and past experience to narrow down potential diseases. However, the complexity and variability of real-world clinical presentations have prompted recent research into computational frameworks that use large language models (LLMs) to improve the DDx process \citep{fansi2022towards,zhou2024interpretable}.
\begin{figure}[t]
	\centering
    \includegraphics[trim={7.2cm 1cm 8cm 3cm},clip,width=0.42\textwidth]{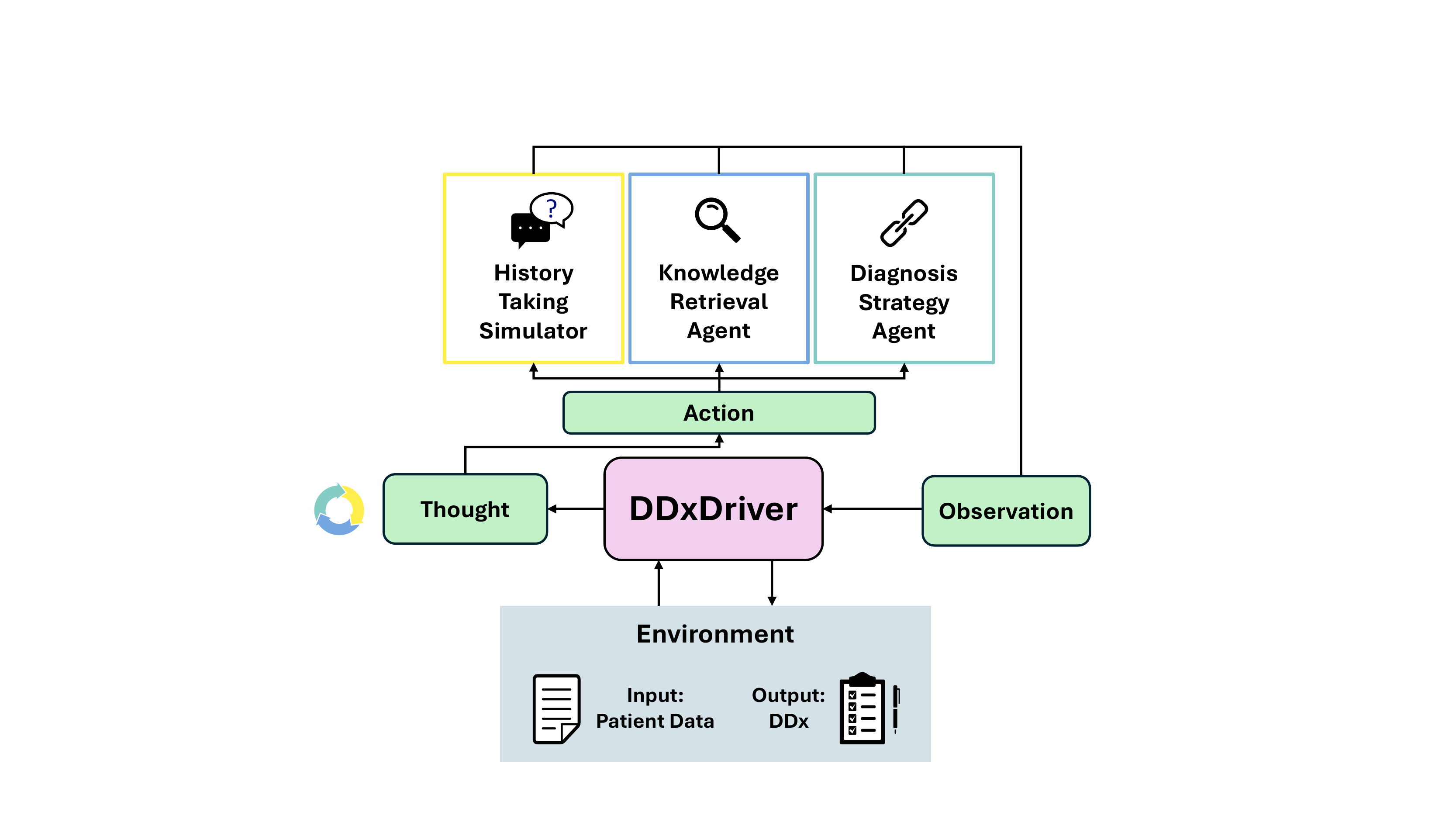}
    \caption{\textbf{MEDDxAgent} facilitates differential diagnosis by iteratively narrowing down a patient's possible disease. DDxDriver acts as the central orchestrator. It receives an interactive environment via a simulator (\textcolor{yellow-history}{\textbf{History Taking}}) and can access two agents (\textcolor{blue-retrieval}{\textbf{Knowledge Retrieval}}, \textcolor{green-diagnosis}{\textbf{Diagnosis Strategy}}).}
    \label{fig:overview_ddxdriver}
\end{figure}

Though LLM-based systems have shown promise in improving diagnostic assistance, existing methods face several limitations: (1) reliance on \textit{single-dataset evaluations}, limiting the generalizability across diverse patient populations and disease categories~\citep{alam2023ddxt}; (2) focus on \textit{optimizing a single diagnostic component} (e.g., diagnosis strategy \textit{only})~\citep{ mcduff2023towards}, without an integrated approach to enhance multiple phases of the diagnostic process;\footnote{The DDx process typically involves three key components: history taking, knowledge retrieval, and diagnosis strategy~\citep{cook2020higher, 2024PrinciplesOD}.} (3) assumption of \textit{complete patient profiles} upfront (i.e., with all symptoms and antecedents)~\citep{wu2024streambench} and \textit{single-turn} paradigm~\citep{zelin2024rare}, diverging from the reality that DDx is an investigative process, requiring follow-up actions to gather information~\citep{li2024mediq}; 
(4) lack of \textit{iterative learning}, preventing diagnosis updates over successive interactions -- an essential aspect of real-world diagnostic decision-making; (5) 
an \textit{over-reliance on medical QA benchmarks}~\citep{zhang2024ultramedical} for medical applications, which do not accurately reflect the complexities of real-world DDx tasks.  
We target these gaps and propose a \textbf{M}odular \textbf{E}xplainable \textbf{DDx} \textbf{Agent} (\textbf{MEDDxAgent}) framework (see~\autoref{fig:overview_ddxdriver}). It consists of (1) 
DDxDriver that acts as the central orchestrator; (2) a history taking simulator which enables an iterative environment; and (3) two individual agents -- knowledge retrieval and diagnosis strategy -- to support the diagnostic process. 
We advance the task of automatic DDx with the following contributions: 
(i) We propose a modular, multi-faceted DDx agent framework (MEDDxAgent), integrating a history taking simulator and two diagnostic agents (knowledge retrieval, diagnosis strategy), which enables extensible and explainable decision-making processes.
(ii) We introduce an orchestrator (DDxDriver) as a unified interface, ensuring \textit{iterative learning} and interactive optimizations between agents, as well as monitoring of the decision-making process. 
(iii) We build a new DDx benchmark incorporating three diagnostic sources with different disease categories: DDxPlus~\citep{fansi2022ddxplus} (\textit{respiratory}), iCRAFT-MD~\citep{li2024mediq} (\textit{skin}), and RareBench~\citep{chen2024rarebench} (\textit{rare}). This allows for a more comprehensive diagnostic scope than in existing work. 
(iv) We evaluate MEDDxAgent in a more challenging but realistic scenario -- \textit{interactive differential diagnosis}, and demonstrate its effectiveness by achieving over 10\% points improvements in accuracy (i.e., GTPA@1) for both large (70B) and small (8B) LLMs.  
(v) The code is publicly available.\footnote{\url{https://github.com/nec-research/meddxagent}}

\section{MEDDxAgent Overview}

\begin{figure*}[t]
	\centering
    \includegraphics[trim={0.5cm 1.0cm 0cm 0cm},clip,width=\textwidth]{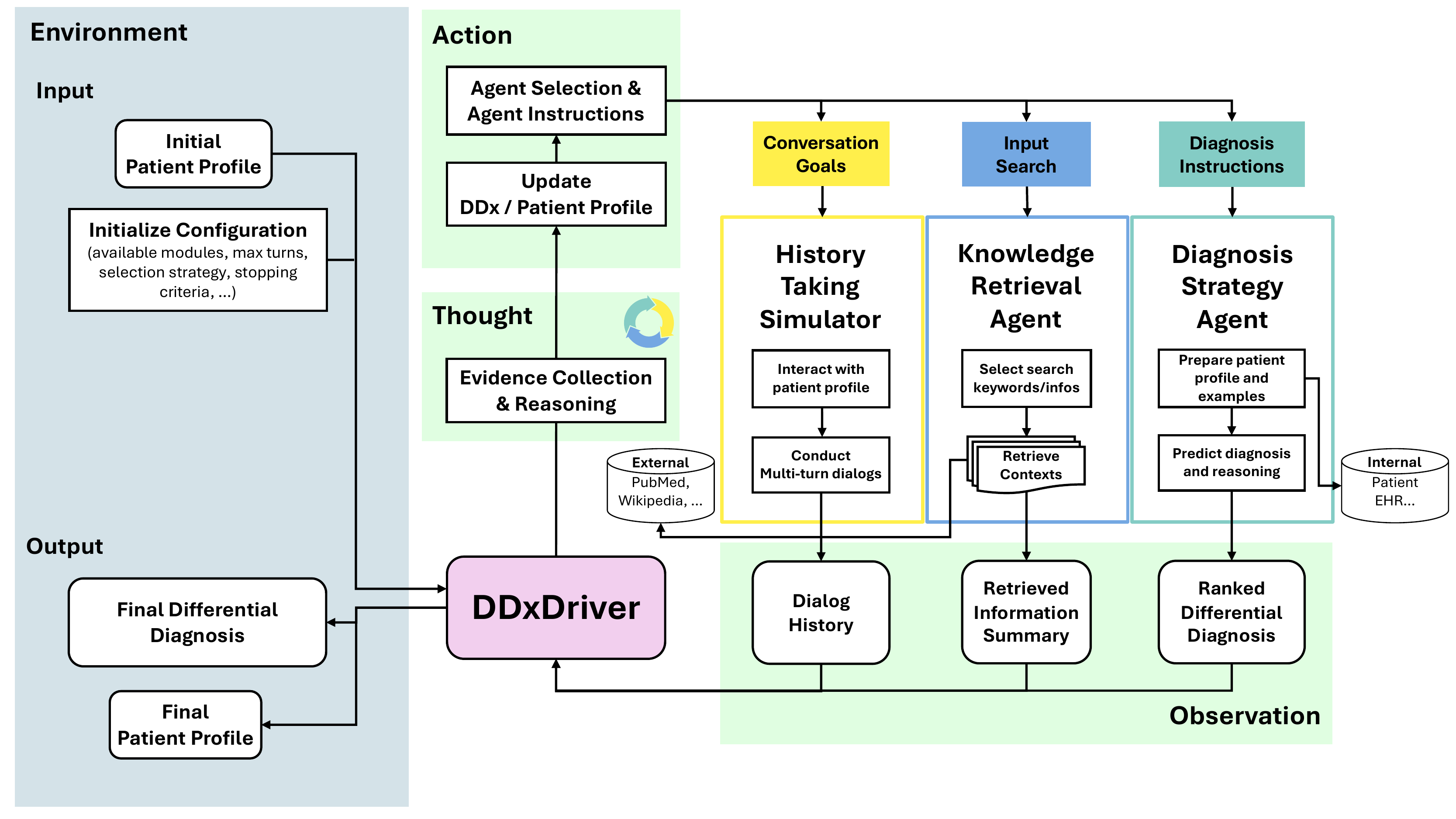}
    \caption{The architecture of the MEDDxAgent framework. MEDDxAgent unifies a central orchestrator (DDxDriver), a simulator (\textcolor{yellow-history}{\textbf{History Taking}}) and two agents (\textcolor{blue-retrieval}{\textbf{Knowledge Retrieval}}, \textcolor{green-diagnosis}{\textbf{Diagnosis Strategy}}). The framework follows the ReAct~\citep{yao2023react} paradigm (\textit{thought}, \textit{action}, \textit{observation}), enabling sequential reasoning and action steps with transparent logging of all interactions through the iterative learning process.}
    \label{fig:overview_framework}
\end{figure*}

Our proposed MEDDxAgent framework (see \autoref{fig:overview_ddxdriver} and a detailed version in~\autoref{fig:overview_framework}) comprises a central orchestrator (DDxDriver), a history taking simulator, and two specialized diagnostic agents dedicated to knowledge retrieval and diagnosis strategy. Both the simulator and diagnostic agents communicate exclusively with the DDxDriver, which monitors, stores, maintains, and updates patient information and ranked differential diagnoses. This central role also positions the DDxDriver to coordinate an iterative feedback loop, wherein observations from each agent are leveraged to enhance and refine subsequent agent calls with agent instructions. In the following, we introduce the design of simulator (\autoref{subsec:simulator}), agents (\autoref{subsec:diagnostic_agents}), orchestrator (DDxDriver) (\autoref{subsec:ddxdriver}), and iterative learning mechanism  (\autoref{subsec:iterative_learning}).




\subsection{Simulator}
\label{subsec:simulator}

History taking is a critical first step in differential diagnosis, where clinicians gather essential information by asking patients questions about their symptoms, medical history, and lifestyle factors. In real-world clinical settings, a full patient profile is rarely available at the outset~\citep{li2024mediq} -- doctors typically start with only partial information (e.g., age, gender, chief complaint). The process of interactive DDx allows clinicians to gather more patient information and refine their diagnostic hypotheses before making a follow-up decision. 

To simulate such an interactive environment, we introduce a history taking simulator. We initialize the simulator with two LLMs \cite{wu2023large} in our experiments. The first LLM simulates the patient and receives access to the full patient profile. The second LLM simulates the doctor and receives an initial patient profile and optionally a set of conversational goals defined by DDxDriver (\textit{action}). During the interactions, the doctor role asks questions relevant to the diagnosis process, and the patient role provides answers based on its patient profile. 
The interaction continues until either the conversational goals are achieved or a predefined stopping criterion (e.g., maximum number of questions) is reached. Once the conversation concludes, the dialogue history is forwarded to DDxDriver.  


\subsection{Agents}
\label{subsec:diagnostic_agents}

\paragraph{Knowledge Retrieval Agent.}
This agent aids the diagnostic process by retrieving relevant medical knowledge from external sources, such as scientific literature, medical databases, and clinical guidelines. This is particularly critical for diagnosing rare or complex conditions where external knowledge (as compared to internal knowledge learned by LLMs' training data) is required to enhance clinical reasoning with validated information.

Upon activation, the agent receives a search query formulated by DDxDriver, based on the current patient profile and provisional DDx list. It extracts the key medical concepts from the query as structured keywords, then conducts a targeted search in external databases. We consider two primary sources: Wikipedia and PubMed\footnote{\url{https://pubmed.ncbi.nlm.nih.gov/}}, with the former providing concise summaries of top-ranked pages, while the latter retrieves abstracts of full-access articles. The retrieved knowledge is synthesized into an evidence-based summary, which ensures that the diagnostic reasoning process has access to up-to-date, relevant medical knowledge.




\paragraph{Diagnosis Strategy Agent.}
This agent is responsible for generating, refining, and ranking possible diagnoses based on the information prepared by DDxDriver. 
There are two distinct modes that can be chosen for the diagnosis strategy agent. First, in the zero-shot setting, the LLM predicts the most probable diagnoses solely based on the current patient. This approach is straightforward but may have limited accuracy for complex or rare conditions. Second, in the few-shot setting, the diagnosis strategy agent utilizes additional patient cases to guide its predictions, enabling more context-aware diagnostic reasoning. We explore two variations. First, in a standard few-shot approach, a fixed set of patient examples are selected as references and provided to the model alongside the current patient profile. Second, the dynamic few-shot approach improves upon this by selecting reference cases based on similarity metrics, ensuring that the most relevant patients are included. Patient similarity is determined using embedding-based retrieval, with two embeddings (i.e., BioClinicalBERT~\citep{alsentzer-etal-2019-publicly}, BGE~\citep{xiao-etal-2024-bge}) evaluated to match patients with similar profiles.

We also integrate Chain-of-Thought (CoT) reasoning~\citep{wei2022chain}, guiding the model to explicitly reason through intermediate clinical steps before predicting a diagnosis. CoT can be combined with both standard and dynamic few-shot approaches, with a stepwise rationale for each diagnosis. Inspired by MedPrompt~\citep{nori2023can}, we extend CoT by incorporating structured, example-driven reasoning, where each reference case includes both a diagnosis and an associated CoT explanation. The integration of CoT enables the system to better handle complex cases with diagnostic uncertainty, such as specific skin diseases in iCRAFT-MD which share common symptoms. 

Once the model completes the diagnostic inference, the ranked list of differential diagnoses is returned to DDxDriver, which further refines or finalizes the diagnosis through iterative updates. By distinguishing between zero-shot and few-shot inference strategies, plus dynamic adaptation through embeddings and reasoning techniques, the diagnosis strategy agent aims to enhance both accuracy and generalizability.



\subsection{Orchestrator} 
\label{subsec:ddxdriver}


Inspired by the concept of a unified interface layer from previous work~\citep{gioacchini-etal-2024-agentquest}, we introduce DDxDriver as the central coordination hub in the MEDDxAgent framework (\autoref{fig:overview_ddxdriver}). DDxDriver enables modular compatibility between the diagnostic agents and benchmark datasets, with minimal adaptation efforts. 
DDxDriver uses the ReAct paradigm~\citep{yao2023react} -- which combines step-by-step reasoning (\textit{thought}) with decision-making (\textit{action}) and feedback processing (\textit{observation}). At each step, DDxDriver obtains the information from the \textit{environment} (input/output) and the results from the previous state of the simulator and agents (\textit{observation}, if it exists), then reasons about the current state of evidence (\textit{thought}) and generates agent-specific instructions (\textit{action}) based on the current state of the patient profile. It dispatches these instructions to the selected simulator/agent, executes, and then updates the patient profile with newly obtained information (\textit{action}). 
Beyond execution management, DDxDriver serves four primary functions. First, it manages the patient profile, storing and maintaining all relevant clinical information, including demographics, medical history, symptoms, and evolving diagnostic rankings. Second, it schedules and dispatches diagnostic actions, dynamically determining which simulator/agent to invoke next based on the evolving diagnostic context. Third, it ensures traceability by logging all interactions, including inputs, outputs, and intermediate reasoning steps, thereby providing transparency in the decision-making process. Finally, it enforces stopping criteria by monitoring diagnostic convergence and applying configurable thresholds, such as the number of iterations or the stabilization of ranked diagnoses.

\begin{figure*}[t]
	\centering
    \includegraphics[trim={0cm 0.9cm 0cm 0cm},clip,width=\textwidth]{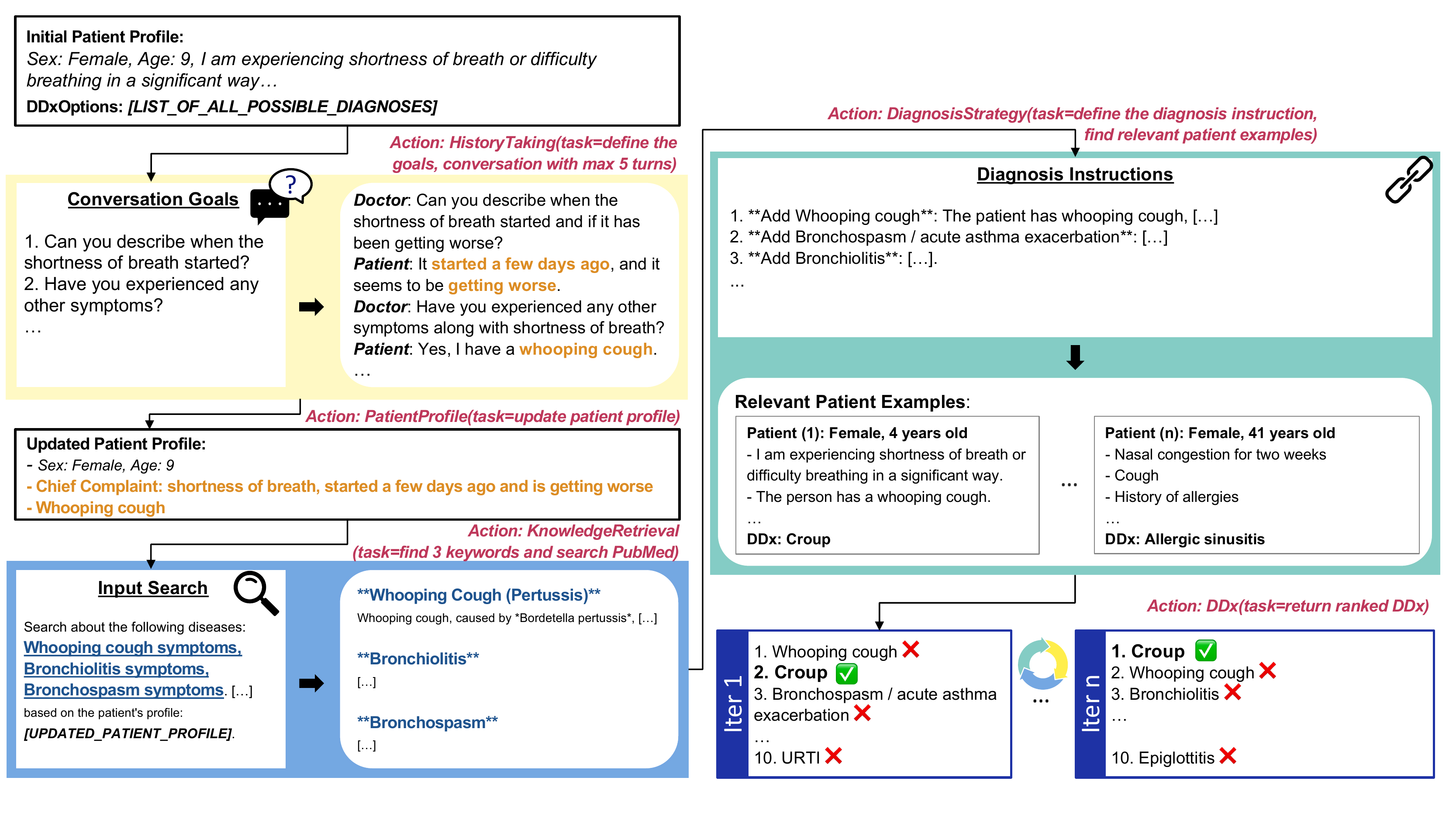}
    \caption{An illustrated DDxPlus~\citep{fansi2022ddxplus} example with the MEDDxAgent framework. Given the initial patient profile and list of diagnosis options, DDxDriver determines the goals and actions for the simulator (\textcolor{yellow-history}{\textbf{History Taking}}) and agents (\textcolor{blue-retrieval}{\textbf{Knowledge Retrieval}}, \textcolor{green-diagnosis}{\textbf{Diagnosis Strategy}}), updating the patient profile, and returning the ranked DDx. Each step is logged for transparency, enabling iterative refinement and learning.} 
    \label{fig:Example}
\end{figure*}

\subsection{Iterative Learning Mechanism}
\label{subsec:iterative_learning}
Diagnoses in the real world are rarely made in a single step. They are refined through multiple interactions with patients, clinical data, and external knowledge. To mirror this process, the \textit{iterative learning} mechanism is designed to avoid relying on any single diagnostic agent or static decision process. 
We implement two settings: (i) \textit{fixed iteration}, and (ii) \textit{dynamic iteration}. \textit{Fixed iteration} cycles through the history taking simulator, knowledge retrieval agent, and diagnosis strategy agent in order until the predefined stopping criterion is met (e.g., $n$ iterations).
In contrast, the \textit{dynamic iteration} process lifts constraints on the predetermined execution order, allowing the DDxDriver to adapt dynamically during the differential diagnosis process. After each observation, the DDxDriver reasons about which component -- history taking simulator, knowledge retrieval agent, or diagnosis strategy agent -- to call next based on up-to-date observations (i.e. updated patient profile, medical documents, predicted DDx). For instance, if the current diagnosis indicates a rare condition for which it needs clarifying details, the system may invoke the knowledge retrieval agent to search for specialized information. Similarly, if the retrieved knowledge introduces ambiguity, the system may loop back to the history taking simulator to clarify new symptoms or risk factors. This allows for flexible decision-making, opening up the opportunity for the diagnostic process to dynamically adjust as new information becomes available. The iterative learning mechanism allows MEDDxAgent to continuously refine diagnosis while offering transparent insights into its reasoning process. A detailed workflow example is illustrated in \autoref{fig:Example}.

\section{Experimental Setup}
\subsection{DDx Benchmark}
\label{subsec:ddx_benchmark}

We introduce a comprehensive DDx benchmark integrating three datasets -- DDxPlus, iCraft-MD, and RareBench, covering \textit{respiratory}, \textit{skin}, and \textit{rare} diseases for a robust assessment of diagnostic performance. This addresses limitations of prior work, which often relies on a single dataset and single-turn evaluation for differential diagnosis. DDxPlus~\citep{fansi2022ddxplus} provides a large-scale, structured dataset with 1.3 million synthetic respiratory patient cases across 49 respiratory-related pathologies. iCraft-MD ~\citep{li2024mediq} includes 394 skin diseases, adapting static dermatological clinical vignettes (from original Craft-MD dataset~\citep{johri2024craft, johri2025craftmd}) into an interactive setting
\footnote{Interactive DDx is a more complex information-seeking setup, since in the real world the full patient profile might not be accessible initially.} -- the system is only provided with partial patient information and is expected to proactively ask questions and gather information. 
RareBench~\citep{chen2024rarebench} expands DDxPlus with 421 rare diseases. We select three subsets from RareBench -- RAMEDIS (Europe), MME (Canada), and PUMCH (China) -- to ensure diversity in regional representation.

To enable a consistent evaluation across datasets, we standardize each dataset into a structured format: (i) optional initial patient information (e.g., age, sex, chief complaint); (ii) full patient profile (complete list of symptoms and antecedents); and (iii) full set of possible diseases for differential diagnosis. 
This refinement enhances diagnostic consistency and supports the evaluation of interactive DDx. We sample 100 patients from each dataset at a fixed random seed, due to the cost of experiments and excessive time for reasoning steps. Detailed dataset statistics are in~\autoref{app:ddx_benchmark_details}.

\subsection{Evaluation Metrics}
To evaluate diagnostic performance, we employ three metrics. First, we compute the \textit{average rank} of the correct disease, which represents the model’s ability to position the correct diagnosis closer to the top. If the diagnosis does not appear in the top-10 position, we assign a rank of 11. Second, we use \textit{GTPA@k} (Ground Truth Pathology Accuracy)~\citep{fansi2022ddxplus}, which measures whether the ground truth diagnosis appears within the top-\textit{k} predicted diagnoses.
Third, we introduce a new metric suitable for the iterative setting: \textit{average progress rate} ($\Delta$ Progress). Inspired by AgentQuest \cite{gioacchini-etal-2024-agentquest}, 
it tracks changes in rank $r$ of the ground truth pathology in the differential diagnosis. For each patient case i, we average the progress in rank ($r_{i,t}-r_{i,t+1}$) over N iterations of differential diagnosis, then aggregate over M patients. This metric quantifies how effectively the system refines and converges on the correct diagnosis over successive iterations:

\small{
\begin{equation*}
\Delta \text{Progress} = \frac{1}{M} \sum_{i=1}^{M} \left( \frac{1}{N_i - 1} \sum_{t=1}^{N_i-1} \Bigl( r_{i,t} - r_{i,t+1} \Bigr) \right)
\end{equation*}}
\normalsize
\subsection{Models and Tasks}

We evaluate on GPT-4o (version: \textsc{2024-11-20})~\citep{hurst2024gpt}, Llama3.1-70B and Llama3.1-8B~\citep{dubey2024llama} across all tasks, ensuring a comparison of LLMs at varying scales.  
Our experiments are conducted in two setups: (1) optimizing individual agents; and (2) interactive differential diagnosis. In the first task, we evaluate the two agents (knowledge retrieval, diagnosis strategy) in a single-turn setting.
This allows us to isolate the effectiveness of the reasoning mechanisms without the confounding factor of incomplete information. In the second task, we 
assess MEDDxAgent's performance at interactive DDx, comparing it against the single-turn diagnostic agents and history taking simulator. Interactive differential diagnosis, as suggested by~\citet{li2024mediq}, is a challenging yet realistic scenario, where only initial patient information is available -- without a complete list of symptoms and antecedents. This setup highlights how limited information constrains the single-turn setting (i.e., no iteration), compared to MEDDxAgent's iterative interactions, which refine and enhance the diagnostic process.


\subsection{Hyperparameters and Optimization}
\label{subsec:hyperaparemeters_optimization}

\setlength{\tabcolsep}{2.3pt}
\begin{table*}[h]
\centering
\scriptsize
\begin{tabular}{rccccccccc}
\toprule
                               & \multicolumn{3}{c}{\textbf{DDxPlus}} & \multicolumn{3}{c}{\textbf{iCraft-MD}} & \multicolumn{3}{c}{\textbf{RareBench}} \\ \cmidrule(lr){2-4} \cmidrule(lr){5-7} \cmidrule(lr){8-10}
                               & \textbf{GTPA@1 $\uparrow$}          & \textbf{GTPA@5 $\uparrow$}   & \textbf{Avg Rank $\downarrow$} & \textbf{GTPA@1 $\uparrow$}       & \textbf{GTPA@5 $\uparrow$}     & \textbf{Avg Rank $\downarrow$}   & \textbf{GTPA@1 $\uparrow$}        & \textbf{GTPA@5 $\uparrow$}   & \textbf{Avg Rank $\downarrow$}     \\\midrule
                               & \multicolumn{9}{c}{\textbf{GPT-4o}}                                                                 \\\midrule
Retrieval (PubMed)                   & 0.69           &  0.90   &  2.27  & 0.68        &     \textbf{0.79}    & 3.23  & 0.45         &   0.72  &    3.92   \\
Retrieval (Wiki)                   &    0.69        &   0.90 &  2.24  & \textbf{0.69}         &     \textbf{0.79}    & \textbf{3.22}  &   0.45       &  0.74  &  4.00    \\ 
\cmidrule(lr){2-10}
Zero-shot (Standard)                     &     0.69           &     0.90       &      2.21      &       0.68         &      0.77        &         3.37       &       0.46       &      0.72       &   3.99             \\
Zero-shot (CoT)                    &     0.71          &     0.92       &      2.10      &       0.68         &     0.77         &         3.35       &       0.47       &    0.69         &   4.02              \\ 
Few-shot (Standard, Dyn\_BAII)$\ddag$ &      0.96          &      \textbf{1.00}      &    1.06         &        0.62        &        0.72      &     3.85           &     0.79         &   \textbf{0.91}          &      \textbf{2.03}           \\
Few-shot (CoT, Dyn\_BERT)      &       0.96         &    \textbf{1.00}        &  1.05        &     0.64           &      0.73        &         3.68      &     0.81         &  \textbf{0.91}            &          2.04      \\
Few-shot (CoT, Dyn\_BAII)      &       \textbf{0.97}         &     \textbf{1.00}       &       \textbf{1.03}      &         0.60       &      0.70        &        4.00      &         \textbf{0.82}     &      0.88       &         2.11      \\

\bottomrule
    \end{tabular}
    \vspace{-0.8em}
    \caption{Results in the non-interactive setting for the knowledge retrieval agent (\textit{upper}) and the diagnosis strategy agent (\textit{bottom}). 
    $\ddag$ Only Few-shot (Standard, Dyn\_BAII) results are recorded, since the method is consistently better than Dyn\_BERT. All models exhibit similar trends. To give a more concise overview, we only report GPT-4o here. The full set of results can be found in~\autoref{tab:with_patient_profile_full} in Appendix.}
    \label{tab:with_patient_profile}
    \vspace{-1.0em}
\end{table*}

For the knowledge retrieval agent, we limit searches to a maximum of three medical keywords per query. Wikipedia is used as an open-access resource, while PubMed retrieval is restricted to full-text articles from commercially licensed sources,\footnote{We use MediaWiki API: \url{https://en.wikipedia.org/w/api.php} and \textsc{biopython}~\url{https://biopython.org/}.} ensuring that retrieved information is clinically validated and relevant to the diagnostic task. For the diagnosis strategy agent, we take 5 examples for few-shot learning. For dynamic few-shot, we use BioClinicalBERT (BERT)~\citep{alsentzer-etal-2019-publicly} and \textsc{bge-base-en-v1.5} (BAII)~\citep{xiao-etal-2024-bge} embeddings, based on the structure proposed by~\citet{wu2024streambench}. Specifically, it uses L2 distance on normalized embeddings, a similar setting to cosine similarity. 
With the history taking simulator, we create an iterative environment, which we evaluate at 5, 10, and 15 maximum questions. This is based on prior clinical studies that indicate physicians typically ask fewer than 15 questions per consultation~\citep{ely1999analysis}. This ensures that our model operates within a realistic range, capturing essential patient details without excessive interaction. 
To evaluate MEDDxAgent's iterative learning, we select the optimized history taking simulator and diagnostic agents and experiment on interactive DDx. Our setup is inspired by previous work~\citep{johri2025craftmd}, which demonstrates that updating the patient profile with new history-taking dialogue significantly enhances performance. We experiment with 1 to 3 iterations, with 5 questions per iteration. This aligns with the history-taking simulator setting (5 questions per iteration, max 15 for 3 iterations). 
Additionally, we set the DDxDriver's instruction for each agent and simulator to a list of length 10.

\section{Evaluation Results}
\label{evaluation-results}

\setlength{\tabcolsep}{3.8pt}
\begin{table*}[ht]
\centering
\scriptsize
\begin{tabular}{rccccccccc}
\toprule
                               & \multicolumn{3}{c}{\textbf{DDxPlus}} & \multicolumn{3}{c}{\textbf{iCraft-MD}} & \multicolumn{3}{c}{\textbf{RareBench}} \\ \cmidrule(lr){2-4} \cmidrule(lr){5-7} \cmidrule(lr){8-10}
                               & \textbf{GTPA@1 $\uparrow$}          & \textbf{Avg Rank $\downarrow$}   & \textbf{$\Delta$ Progress} & \textbf{GTPA@1 $\uparrow$}       & \textbf{Avg Rank $\downarrow$}     & \textbf{$\Delta$ Progress}   & \textbf{GTPA@1 $\uparrow$}        & \textbf{Avg Rank $\downarrow$}   & \textbf{$\Delta$ Progress}     \\\midrule
                               & \multicolumn{9}{c}{\textbf{GPT-4o}}                                                                 \\\midrule
KR (\textit{n}=0)                &      0.18      & 7.33  &  -  &   0.15      &    8.27     & -  &       0.07   &  9.07  &    -   \\
DS (\textit{n}=0)     &  0.27    &    6.01        &       -      &      0.18          &      7.87        &          -      &       0.11       &     8.38        &           -      \\
KR (\textit{n}=5)  &      0.52      & 3.32   &  -  &  0.49  &  5.36       & -  &     0.40   &   5.27 &    -   \\
DS (\textit{n}=5)  &    0.72       &  2.14 &  -  &  0.40 &    5.55   & -  &   0.50    &   4.94 &    -   \\\cmidrule(lr){2-10}

 MEDDx (\textit{iter}=1, \textit{n}=5)                       & 0.74           & 1.91     & ~~0.00 & 0.52        & 4.93      &  ~~0.00   & 0.51         & 4.37        &   ~~0.00\\
MEDDx (\textit{iter}=2, \textit{n}=10)                       & 0.78           & 1.56    & +0.32  & \textbf{0.54}        & \textbf{4.71}    &    +0.26   & \textbf{0.56}         & 4.10   &     +0.13   \\
MEDDx (\textit{iter}=3, \textit{n}=15)                       & \textbf{0.86}           & \textbf{1.29}    & +0.32  & \textbf{0.54}        & 4.80      & +0.17    & 0.50         & \textbf{4.09}       &   +0.16 \\\midrule
                               & \multicolumn{9}{c}{\textbf{Llama3.1-70B}}                                                           \\ \midrule
KR (\textit{n}=0)           &   0.19         &  7.58  &  -  &      0.13   &   8.19      & -  &    0.09      &  9.13  &    -   \\
DS (\textit{n}=0)    &        0.17        &       7.28     &       -      &      0.11          &      8.74        &          -      &       0.20       &      6.81       &           -      \\
KR (\textit{n}=5)  &      0.39      & 5.03   &  -  &  0.34  &  6.86       & -  &     0.29   &   5.86 &    -   \\
DS (\textit{n}=5)  &     0.50      &  2.89 &  -  & 0.24 &    7.33   & -  &   0.23    &  5.77  &    -   \\\cmidrule(lr){2-10}
MEDDx (\textit{iter}=1, \textit{n}=5)                       & 0.61           & 2.91    & ~~0.00  & 0.29        & 7.05       & ~~0.00   & 0.39         & 5.05   &     ~~0.00   \\
MEDDx (\textit{iter}=2, \textit{n}=10)                       & \textbf{0.71}   & \textbf{2.20}      & +0.41 & 0.37        & \textbf{6.26}      & +0.07   & \textbf{0.48}         & 4.48  &    +0.75     \\
MEDDx (\textit{iter}=3, \textit{n}=15)                       & 0.68   & 2.30    & +0.17  & \textbf{0.42}        & 6.31     &   +0.26   & \textbf{0.48}         & \textbf{4.30}      &   +0.44  \\\midrule
                               & \multicolumn{9}{c}{\textbf{Llama3.1-8B}}                                                            \\\midrule
KR (\textit{n}=0)     &     0.20       &  7.49  &  -  &   0.11      &  \textbf{8.86}       & -  &     \textbf{0.11}     &  8.58  &    -   \\
DS (\textit{n}=0)  &       0.16         &       8.45     &       -      &      0.03          &      10.37        &          -      &         0.04     &       8.52      &           -      \\
KR (\textit{n}=5)  &      0.21      & 7.42   &  -  &  0.09  &  9.48       & -  &     0.04   &   9.69 &    -   \\
DS (\textit{n}=5)  &     0.23      &  5.77  &  -  &  0.03 &   10.08    & -  &   0.06    &  8.64  &    -   \\\cmidrule(lr){2-10}
MEDDx (\textit{iter}=1, \textit{n}=5)                       & 0.34           & 5.25   &   ~~0.00 & 0.11        & 9.38       &  ~~0.00  & 0.08         & 8.47    &    ~~0.00   \\
MEDDx (\textit{iter}=2, \textit{n}=10)                       & 0.56           & 3.59    & +1.73  & \textbf{0.14}        & 9.22       &  +0.22  & 0.09         & \textbf{8.11}      &  +0.44   \\
MEDDx (\textit{iter}=3, \textit{n}=15)                       & \textbf{0.58}           & \textbf{3.10}    &  +1.23 & 0.12        & 9.07     & +0.17     & 0.07         & 8.56    &  +0.38  \\  
\bottomrule
    \end{tabular}
    \vspace{-0.8em}
    \caption{Interactive experiment performance across 3 datasets without \textit{full} patient profile, with KR: knowledge retrieval agent; DS: diagnosis strategy agent; $n$ is the number of turns of the simulator; MEDDx uses KR+DS.
    }
    \label{tab:interactive_overall}
    \vspace{-1.8em}
\end{table*}

We experiment on two configurations: (1) optimizing individual agents (\autoref{subsec:optimize-agents}), by determining the best settings for knowledge retrieval and diagnosis strategy agents; and (2) interactive differential diagnosis (\autoref{subsec:iterative_learning}), where the optimized agents are used to assess MEDDxAgent's performance in the interactive DDx setup.

\subsection{Optimizing Individual Agents}
\label{subsec:optimize-agents}

We first explore the optimal single-turn configuration for the knowledge retrieval and diagnosis strategy agents, before integrating them into the iterative setup. For this, we provide the full patient profile as in previous work~\cite{wu2024streambench,chen2024rarebench}, and present the results in~\autoref{tab:with_patient_profile}. 
For the knowledge retrieval agent, PubMed performs slightly better overall than Wikipedia, especially for Rarebench, which demands more complex disease information. 
For the diagnosis strategy agent, the best setting varies by dataset. 
Namely, dynamic few-shot with BAII embeddings performs the best on DDxPlus and RareBench, where relevant patient examples offer reliable contextual cues to likely diseases. 
In contrast, iCraft-MD benefits more from zero-shot CoT, which enables structured reasoning through complex clinical vignettes. Few-shot learning often decreases performance for iCraft-MD because each patient vignette is distinct, so additional examples can introduce noise.
Based on the above findings, we select the following configurations for the iterative scenario:\footnote{We do not run all possible settings in the interactive environment due to cost reasons.} PubMed for knowledge retrieval agent; few-shot (dynamic BAII) for DDxPlus and RareBench, and zero-shot (CoT) for iCraft-MD for diagnosis strategy agent.

\begin{figure*}[t]
    \centering
    \begin{subfigure}{0.48\textwidth}
    \includegraphics[trim={0.2cm 0cm 0cm 0cm },clip, width=\textwidth]{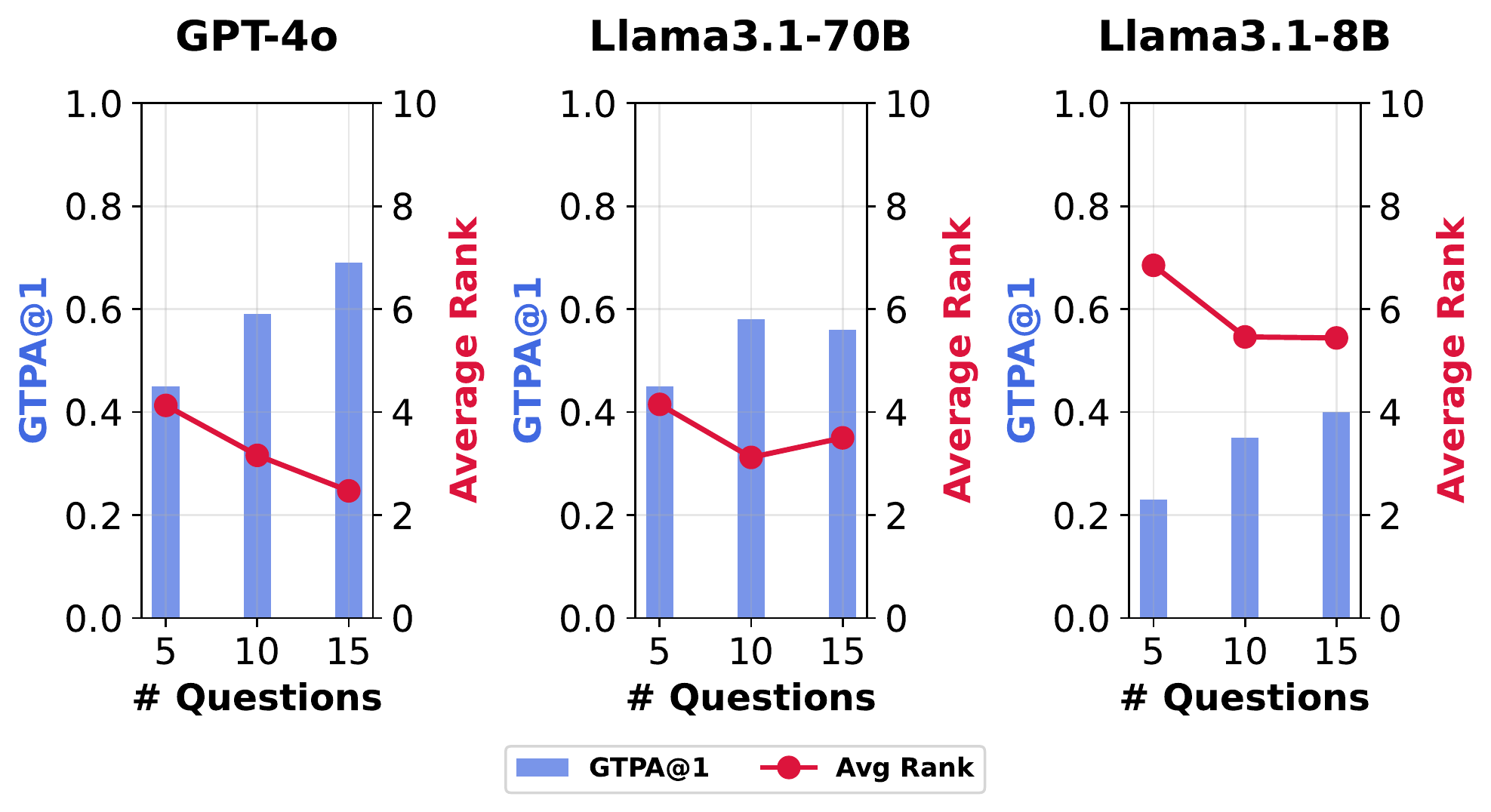}
    \caption{}
    \end{subfigure}
    \begin{subfigure}{0.48\textwidth}
    \includegraphics[trim={0.2cm 0cm 0cm 0cm}, clip, width=\textwidth]{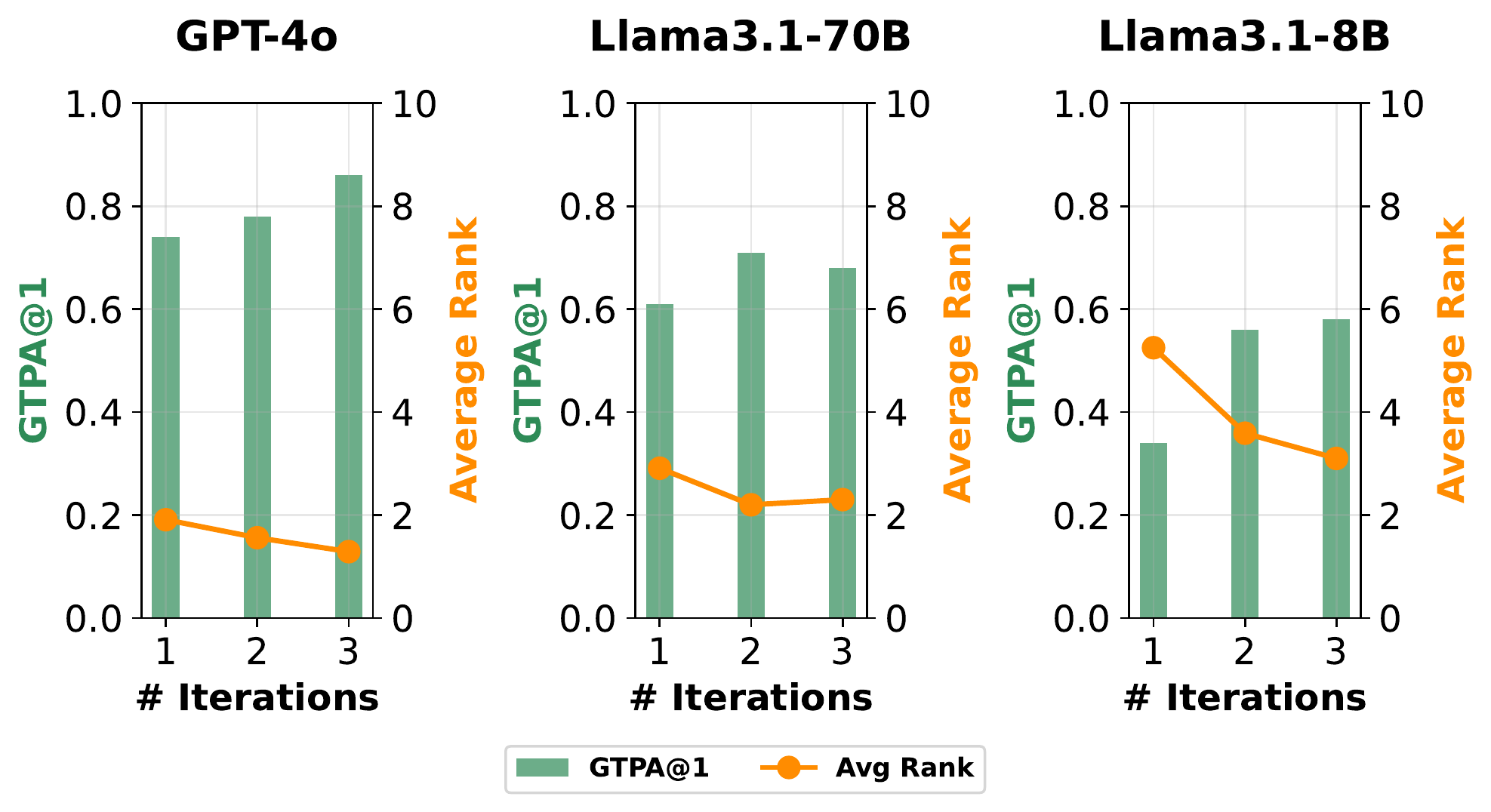}
    \caption{}
    \end{subfigure}
    \caption{Results of DDxPlus compared between (a) history taking simulator, and (b) MEDDxAgent, over the number of questions and iterations. For brevity, the results of iCraft-MD and RareBench are in~\autoref{subsec:comparison_history_taking_iterative}.}
    \label{fig:ddxplus_comparison}
\end{figure*}

\subsection{Interactive Differential Diagnosis}
\label{subsec:interactive_differential_diagnosis}
We now evaluate the more challenging task of interactive DDx, where we begin with limited patient information and the history taking simulator enables the interactive environment~(\autoref{tab:interactive_overall}).
The process is initialized with the number of turns $n$ of the history taking simulator, coupled with either (a) knowledge retrieval agent only (KR), (b) diagnosis strategy agent only (DS), or (c) MEDDxAgent with both knowledge retrieval and diagnosis agents (MEDDx).  
At $n=0$, the simulator has not yet learned any patient information, and performance drops significantly from observing the full patient profile (\autoref{tab:with_patient_profile}). 
For GPT-4o in RareBench, the knowledge retrieval agent (KR)'s GTPA@1 drops from 0.45  to 0.07. Similarly, the diagnosis strategy agent (DS) drops from 0.46 (zero-shot) to 0.11. This simple baseline showcases that previous evaluations do not hold well in the interactive setup with initially limited patient information. 
Already for $n=5$, we find a large boost in performance for both KR and DS. These findings emphasize the importance of history taking for diagnostic precision. 
We illustrate the trend for changing $n$ in~\autoref{fig:ddxplus_comparison} and find that gains also plateau around \textit{n}=10-15 questions, reinforcing the optimal balance between information gathering and diagnostic efficiency \cite{ely1999analysis}.


Finally, we run MEDDxAgent, which calls KR+DS in the \textit{fixed iteration} pipeline (\autoref{subsec:iterative_learning}).
MEDDxAgent exhibits clear improvements over the KR and DS baselines for $n=5$, supporting our hypothesis that all three modules are important for interactive DDx. It also improves significantly over the history taking baselines, as we illustrate in \autoref{fig:ddxplus_comparison}. MEDDxAgent is also capable of improving upon the zero-shot setting with the full patient profile (\autoref{tab:with_patient_profile}). For DDxPlus, GTPA@1 for GPT-4o and Llama3.1-70B rise from 0.69 to 0.86 and from 0.54 to 0.71, respectively. For Llama3.1-8B, the trend continues for DDxPlus but inconsistently for iCraft-MD and RareBench, highlighting the importance of model scale. Notably, MEDDxAgent improves over successive iterations, though the optimal number of iterations (2, 3) depends on the dataset and LLM. The values of $\Delta$ are consistently positive, indicating that MEDDxAgent iteratively increases the rank of the ground-truth diagnosis over time. $\Delta$ Progress also varies by dataset and model, offering explainable insight to the diagnosistic improvement of MEDDxAgent. The overall results show that MEDDxAgent can operate well in the challenging, realistic setup of interactive DDx. Additionally, MEDDxAgent logs all intermediate reasoning, action, and observations, providing critical insight into its DDx process (\autoref{fig:Example}).

\section{Analysis}
\label{analysis}
\paragraph{Fixed vs. Dynamic Iterations.}
A key feature of MEDDxAgent is its iterative DDx process, which operates in \textit{fixed} or \textit{dynamic} iteration. Our experiments (see~\autoref{tab:choice_merged} in Appendix) show that fixed iteration consistently outperforms dynamic iteration in both accuracy and system efficiency.\footnote{On average, fixed iteration is 1.2x-1.7x faster than dynamic iteration.} Fixed iteration ensures a structured sequence where all modules -- history taking simulator, knowledge retrieval agent, and diagnosis strategy agent, are utilized in each cycle, preventing over-reliance on a single component. In contrast, dynamic iteration, which allows DDxDriver to choose the component at each step, introduces some suboptimal decision-making. We observe that Llama3.1-* models, for instance, frequently favor the history taking simulator rather than leveraging the knowledge retrieval or diagnosis strategy agents, leading to redundant questioning rather than efficient diagnostic reasoning. Despite this, our findings demonstrate the general applicability of MEDDxAgent for dynamic iteration and highlight future work toward optimizing dynamic iteration for interactive DDx.

\paragraph{Error Analysis.}
To better understand the struggles of MEDDxAgent, we conduct error analysis on cases where it failed to reach the correct diagnosis efficiently. We emphasize that our MEDDxAgent's logging of intermediate logic greatly enhances our understanding and explanations of failure cases. First, in RareBench, over-reliance on few-shot examples often misprioritizes frequent conditions over rarer diseases, as some rare conditions are underrepresented in knowledge retrieval databases. 
Second, in the first iteration, MEDDxAgent tends to prioritize the knowledge retrieval while overlooking few-shot patient examples with similar profiles. This is then mitigated when further iterations help to refine and enhance the DDx process.
Third, larger models (GPT-4o, Llama3.1-70B) benefit more from iterative refinement, while smaller models (Llama3.1-8B) plateau after the second iteration, especially for iCraft-MD and RareBench. Fourth, recent studies have shown that domain-specific medical LLMs often underperform general-purpose models in diagnostic tasks \cite{nori2023can, nori2023capabilities, maharjan2024openmedlm}. We hypothesize this occurs because such models are typically fine-tuned on specific tasks or disease specialities, limiting their general instruction-following capability and diagnostic accuracy. Our experiments confirm this hypothesis, demonstrating that medical LLMs fine-tuned on tasks such as QA and open-ended medical chat \cite{zhang2024ultramedical} perform worse than general LLMs of similar sizes (see \autoref{app:additional_experiments}). Addressing these challenges can further enhance MEDDxAgent’s diagnostic accuracy.

\section{Related Work}
\subsection{LLM-based Methods} 
\label{subsec:llm-based-methods}
Researchers have studied the capabilities of LLMs for automatic diagnosis \cite{mizuta2024diagnosis}. One line of work has found that the performance of LLMs is comparable to the performance of physicians \cite{hirosawa2023chatgpt,rutledge2024diagnostic}, or that the performance of the physicians themselves is improved when they use LLMs \cite{ten2024chatgpt}. However, researchers have also observed that LLMs struggle to perform this task when applied on rarer diseases or on more unusual cases \cite{fabre2024evaluating,shikino2024evaluation}. For this reason we assemble a benchmark that measures performance for different rarity levels.

Many of the current methods are typically targeting either automatic standard diagnosis or automatic differential diagnosis in an end-to-end manner. For example, there are methods that use Chain-of-Thought strategies \cite{wu2023large,savage2024diagnostic,nachane2024few}, reinformcement learning \cite{fansi2022towards}, fine-tuning LLMs \cite{alam2023ddxt, saab2024capabilities, reese2024limitations}, preranking-reranking methods \cite{sun2024conversational} or specifically trained neural networks \cite{liu2020deep,hwang2022differential}. 
Such LLM-based methods do not allow for modularity, posing difficulties for integrating specific modules that solve sub-problems within the diagnosis process. 

\subsection{Agent-based Methods}


Recent work has shifted from standalone LLMs to multi-agent frameworks, enhancing efficiency by enabling external tools, assigning specialized roles to each agent to accomplish complex tasks more efficiently. In medical applications, agent-based methods streamline clinical workflows to improve diagnostic accuracy. KG4Diagnosis~\citep{zuo2024kg4diagnosis} integrated LLMs with knowledge graphs (KGs) for medical diagnosis. However, its static KG dependence makes expanding diagnoses for rare diseases difficult, and the lack of iterative refinement limits adaptability to evolving clinical cases. 
\citet{wu2024streambench} presented StreamBench to evaluate the continuous improvement of LLM agents in streaming environments via simulated feedback. However, it evaluates on full patient profiles and a single DDx dataset, limiting its generalizability. Among recent advances, multi-turn diagnostic frameworks such as CoD~\citep{chen2024cod} and MediQ~\citep{li2024mediq} have placed particular emphasis on modeling the history-taking process through iterative dialogue. In these systems, agents engage in multi-turn interactions to simulate how clinicians gather patient history: asking follow-up questions and incrementally collecting information needed for differential diagnosis. The focus on iterative, dialogue-based history taking brings agent-based systems closer to real-world clinical reasoning, where uncertainty is managed through successive inquiry and response. Other frameworks, such as AMIE~\citep{tu2024towards}, AMSC~\cite{wang2024beyond}, AgentHospital \cite{li2024agent}, MedAgents~\citep{tang-etal-2024-medagents} and MedAgentBench \cite{jiang2025medagentbench}, address different aspects of clinical interactions. However, these approaches suffer from (i) lack of iterative refinement, relying on single-turn reasoning; (ii) limited evaluation, focusing on a single dataset or a single diagnostic component, limiting diagnostic generalization; (iii) assuming full patient profiles as the input, which does not reflect real-world interactive differential diagnosis in which clinicians collect information iteratively. We address these aspects in our work, proposing a more challenging interactive setting and a modular, iterative agent framework. 
\section{Conclusions}
Existing approaches to automatic differential diagnosis rely on single-dataset evaluations, assume fully observed patient profiles, focus on optimizing isolated diagnostic components, or diagnose in a single attempt. We introduce MEDDxAgent, a modular and explainable framework that enhances automatic differential diagnosis through iterative learning. MEDDxAgent integrates a history taking simulator, two agents (knowledge retrieval, diagnosis strategy), and an orchestrator (DDxDriver) to tackle the more challenging and realistic scenario of interactive differential diagnosis, where patient profiles are initially incomplete. Its modular design enables systematic evaluation of optimal agent configurations, while intermediate logging and a novel average progress rate metric provide critical transparency into its reasoning process. Experimental results demonstrate that interactive differential diagnosis is significantly more challenging, allowing MEDDxAgent to iteratively refine predictions and outperform simpler, single-turn approaches. We hope this framework fosters continued progress in developing more adaptive and effective models for automatic differential diagnosis.

\section*{Acknowledgments}
We would like to thank Andreas Ripke for the support of the infrastructure and the anonymous reviewers for their valuable feedback.

\section*{Limitations}
While our proposed MEDDxAgent framework advances the task of automatic differential diagnosis through a modular, explainable, and interactive approach, we acknowledge certain limitations. (1) Model Selection: Our evaluation focuses on Llama models (8B and 70B) and GPT-4o. These models demonstrate strong instruction-following capabilities, so findings may not generalize to all LLM architectures. We also evaluate state-of-the-art medical-domain LLMs~\citep{zhang2024ultramedical} but find that these instruction-tuned models underperform the general-purpose LLMs (e.g., Llama3.1~\citep{dubey2024llama}), likely due to diverse instruction-following behavior. Further exploration is needed for models with different architectures, training paradigms, or domain adaptations. (2) Language Coverage: our framework is evaluated primarily on an English DDx benchmark, limiting its applicability to non-English-speaking regions, where medical terminology, case presentations, and healthcare practices may differ significantly. Extending the framework for multilingual and cross-lingual diagnostic tasks remains an important direction \cite{qiu2024towards, garcia-ferrero-etal-2024-medmt5}. (3) Multimodality: Medical information often relies on multimodal data, such as medical imaging and videos, laboratory tests, electronic health records, and genomic/pathology data. Our DDx Benchmark is soley text-based, which limits the scope of its applicability to clinical decision making. Future work should explore multimodal agentic frameworks \cite{schmidgall2024agentclinic, kim2024mdagents} and reasoning over multimodal sequential data \cite{rose2023visual, himakunthala-etal-2023-lets, zhu2024emerge}. (4) Benchmark Dataset Selection: We introduce a DDx benchmark encompassing respiratory, skin, and rare diseases. Though our benchmark offers greater diversity than prior work, it does not cover all medical specialties or real-world patient distributions. Expanding datasets to reflect a wider range of diseases, demographics, and clinical settings would improve generalizability. (5) MEDDxAgent requires significant communication among components and tools (i.e., DDxDriver, agents, patient profiles, web search, ...), allowing higher latency and computational cost for more thorough reasoning and improved diagnostic accuracy. Future deployments could reduce latency through strategies like parallelism, caching, memory optimization, model selection, and prompt engineering. (6) Training and Deployment: Although MEDDxAgent provides all intermediate reasoning logs and citations, proper physician training is essential to avoid over-reliance on AI-generated diagnoses, which may contain inaccuracies or hallucinations.

Despite these limitations, we hope this work contributes to advancing automatic DDx with a challenging yet realistic setup -- interactive differential diagnosis. We hope that future research builds on top of our findings to include more languages, modalities, datasets, and a deeper exploration of diagnostic processes to enhance the applicability and effectiveness of interactive diagnostic models.


\bibliography{custom}

\begin{thebibliography}{56}
\providecommand{\natexlab}[1]{#1}

\bibitem[{Alam et~al.(2023)Alam, Raff, Oates, and Matuszek}]{alam2023ddxt}
Mohammad~Mahmudul Alam, Edward Raff, Tim Oates, and Cynthia Matuszek. 2023.
\newblock \href {https://openreview.net/forum?id=Uk6WMt9l9w} {{DD}xt: Deep generative transformer models for differential diagnosis}.
\newblock In \emph{Deep Generative Models for Health Workshop NeurIPS 2023}.

\bibitem[{Alsentzer et~al.(2019)Alsentzer, Murphy, Boag, Weng, Jindi, Naumann, and McDermott}]{alsentzer-etal-2019-publicly}
Emily Alsentzer, John Murphy, William Boag, Wei-Hung Weng, Di~Jindi, Tristan Naumann, and Matthew McDermott. 2019.
\newblock \href {https://doi.org/10.18653/v1/W19-1909} {Publicly available clinical {BERT} embeddings}.
\newblock In \emph{Proceedings of the 2nd Clinical Natural Language Processing Workshop}, pages 72--78, Minneapolis, Minnesota, USA. Association for Computational Linguistics.

\bibitem[{Chen et~al.(2024{\natexlab{a}})Chen, Gui, Gao, Ji, Wang, Wan, and Wang}]{chen2024cod}
Junying Chen, Chi Gui, Anningzhe Gao, Ke~Ji, Xidong Wang, Xiang Wan, and Benyou Wang. 2024{\natexlab{a}}.
\newblock \href {https://arxiv.org/pdf/2407.13301} {Cod, towards an interpretable medical agent using chain of diagnosis}.
\newblock \emph{arXiv preprint arXiv:2407.13301}.

\bibitem[{Chen et~al.(2024{\natexlab{b}})Chen, Mao, Guo, Wang, Zhang, and Chen}]{chen2024rarebench}
Xuanzhong Chen, Xiaohao Mao, Qihan Guo, Lun Wang, Shuyang Zhang, and Ting Chen. 2024{\natexlab{b}}.
\newblock \href {https://doi.org/10.1145/3637528.3671576} {Rarebench: Can llms serve as rare diseases specialists?}
\newblock In \emph{Proceedings of the 30th ACM SIGKDD Conference on Knowledge Discovery and Data Mining}, KDD '24, page 4850–4861, New York, NY, USA. Association for Computing Machinery.

\bibitem[{Cook and Décary(2020)}]{cook2020higher}
Chad~E. Cook and Simon Décary. 2020.
\newblock \href {https://doi.org/10.1016/j.bjpt.2019.01.010} {Higher order thinking about differential diagnosis}.
\newblock \emph{Brazilian Journal of Physical Therapy}, 24(1):1--7.

\bibitem[{Dubey et~al.(2024)Dubey, Jauhri, Pandey, Kadian, Al-Dahle, Letman, Mathur, Schelten, Yang, Fan et~al.}]{dubey2024llama}
Abhimanyu Dubey, Abhinav Jauhri, Abhinav Pandey, Abhishek Kadian, Ahmad Al-Dahle, Aiesha Letman, Akhil Mathur, Alan Schelten, Amy Yang, Angela Fan, et~al. 2024.
\newblock \href {https://arxiv.org/pdf/2407.21783} {The llama 3 herd of models}.
\newblock \emph{arXiv preprint arXiv:2407.21783}.

\bibitem[{Ely et~al.(1999)Ely, Osheroff, Ebell, Bergus, Levy, Chambliss, and Evans}]{ely1999analysis}
John~W Ely, Jerome~A Osheroff, Mark~H Ebell, George~R Bergus, Barcey~T Levy, M~Lee Chambliss, and Eric~R Evans. 1999.
\newblock \href {https://www.bmj.com/content/319/7206/358.1.full} {Analysis of questions asked by family doctors regarding patient care}.
\newblock \emph{Bmj}, 319(7206):358--361.

\bibitem[{Fabre et~al.(2024)Fabre, {Magalhaes Filho}, Aguiar, {da Costa}, Gutierres, William, and {Del Giglio}}]{fabre2024evaluating}
B.L. Fabre, M.A.F. {Magalhaes Filho}, P.N. Aguiar, F.M. {da Costa}, B.~Gutierres, W.N. William, and A.~{Del Giglio}. 2024.
\newblock \href {https://doi.org/10.1016/j.esmorw.2024.100042} {Evaluating gpt-4 as an academic support tool for clinicians: a comparative analysis of case records from the literature}.
\newblock \emph{ESMO Real World Data and Digital Oncology}, 4:100042.

\bibitem[{Fansi~Tchango et~al.(2022{\natexlab{a}})Fansi~Tchango, Goel, Martel, Wen, Marceau~Caron, and Ghosn}]{fansi2022towards}
Arsene Fansi~Tchango, Rishab Goel, Julien Martel, Zhi Wen, Gaetan Marceau~Caron, and Joumana Ghosn. 2022{\natexlab{a}}.
\newblock \href {https://proceedings.neurips.cc/paper_files/paper/2022/file/9b6c8c4a5aeb6a37c9efa963e30993d9-Paper-Conference.pdf} {Towards trustworthy automatic diagnosis systems by emulating doctors' reasoning with deep reinforcement learning}.
\newblock \emph{Advances in Neural Information Processing Systems (NeurIPS)}, 35:24502--24515.

\bibitem[{Fansi~Tchango et~al.(2022{\natexlab{b}})Fansi~Tchango, Goel, Wen, Martel, and Ghosn}]{fansi2022ddxplus}
Arsene Fansi~Tchango, Rishab Goel, Zhi Wen, Julien Martel, and Joumana Ghosn. 2022{\natexlab{b}}.
\newblock \href {https://proceedings.neurips.cc/paper_files/paper/2022/file/cae73a974390c0edd95ae7aeae09139c-Paper-Datasets_and_Benchmarks.pdf} {Ddxplus: A new dataset for automatic medical diagnosis}.
\newblock In \emph{Advances in Neural Information Processing Systems}, volume~35, pages 31306--31318.

\bibitem[{Garc{\'i}a-Ferrero et~al.(2024)Garc{\'i}a-Ferrero, Agerri, Atutxa~Salazar, Cabrio, de~la Iglesia, Lavelli, Magnini, Molinet, Ramirez-Romero, Rigau, Villa-Gonzalez, Villata, and Zaninello}]{garcia-ferrero-etal-2024-medmt5}
Iker Garc{\'i}a-Ferrero, Rodrigo Agerri, Aitziber Atutxa~Salazar, Elena Cabrio, Iker de~la Iglesia, Alberto Lavelli, Bernardo Magnini, Benjamin Molinet, Johana Ramirez-Romero, German Rigau, Jose~Maria Villa-Gonzalez, Serena Villata, and Andrea Zaninello. 2024.
\newblock \href {https://aclanthology.org/2024.lrec-main.974/} {{M}ed{MT}5: An open-source multilingual text-to-text {LLM} for the medical domain}.
\newblock In \emph{Proceedings of the 2024 Joint International Conference on Computational Linguistics, Language Resources and Evaluation (LREC-COLING 2024)}, pages 11165--11177, Torino, Italia. ELRA and ICCL.

\bibitem[{Gioacchini et~al.(2024)Gioacchini, Siracusano, Sanvito, Gashteovski, Friede, Bifulco, and Lawrence}]{gioacchini-etal-2024-agentquest}
Luca Gioacchini, Giuseppe Siracusano, Davide Sanvito, Kiril Gashteovski, David Friede, Roberto Bifulco, and Carolin Lawrence. 2024.
\newblock \href {https://doi.org/10.18653/v1/2024.naacl-demo.19} {{A}gent{Q}uest: A modular benchmark framework to measure progress and improve {LLM} agents}.
\newblock In \emph{Proceedings of the 2024 Conference of the North American Chapter of the Association for Computational Linguistics: Human Language Technologies (Volume 3: System Demonstrations)}, pages 185--193, Mexico City, Mexico. Association for Computational Linguistics.

\bibitem[{Henderson et~al.(2012)Henderson, Tierney~Jr, and Smetana}]{henderson2012patient}
Mark~C Henderson, Lawrence~M Tierney~Jr, and Gerald~W Smetana. 2012.
\newblock The patient history: An evidence-based approach to differential diagnosis.
\newblock \emph{McGraw Hill Professional}.

\bibitem[{Himakunthala et~al.(2023)Himakunthala, Ouyang, Rose, He, Mei, Lu, Sonar, Saxon, and Wang}]{himakunthala-etal-2023-lets}
Vaishnavi Himakunthala, Andy Ouyang, Daniel Rose, Ryan He, Alex Mei, Yujie Lu, Chinmay Sonar, Michael Saxon, and William Wang. 2023.
\newblock \href {https://doi.org/10.18653/v1/2023.emnlp-main.15} {Let`s think frame by frame with {VIP}: A video infilling and prediction dataset for evaluating video chain-of-thought}.
\newblock In \emph{Proceedings of the 2023 Conference on Empirical Methods in Natural Language Processing}, pages 204--219, Singapore. Association for Computational Linguistics.

\bibitem[{Hirosawa et~al.(2023)Hirosawa, Kawamura, Harada, Mizuta, Tokumasu, Kaji, Suzuki, and Shimizu}]{hirosawa2023chatgpt}
Takanobu Hirosawa, Ren Kawamura, Yukinori Harada, Kazuya Mizuta, Kazuki Tokumasu, Yuki Kaji, Tomoharu Suzuki, and Taro Shimizu. 2023.
\newblock \href {https://medinform.jmir.org/2023/1/e48808} {{ChatGPT-Generated Differential Diagnosis Lists for Complex Case–Derived Clinical Vignettes: Diagnostic Accuracy Evaluation}}.
\newblock \emph{JMIR Medical Informatics}, 11.

\bibitem[{Hurst et~al.(2024)Hurst, Lerer, Goucher, Perelman, Ramesh, Clark, Ostrow, Welihinda, Hayes, Radford et~al.}]{hurst2024gpt}
Aaron Hurst, Adam Lerer, Adam~P Goucher, Adam Perelman, Aditya Ramesh, Aidan Clark, AJ~Ostrow, Akila Welihinda, Alan Hayes, Alec Radford, et~al. 2024.
\newblock \href {https://arxiv.org/pdf/2410.21276?} {Gpt-4o system card}.
\newblock \emph{arXiv preprint arXiv:2410.21276}.

\bibitem[{Hwang et~al.(2022)Hwang, Choi, Choi, Ju, Kim, Hong, Lee, Yoon, Park, Lee et~al.}]{hwang2022differential}
In-Chang Hwang, Dongjun Choi, You-Jung Choi, Lia Ju, Myeongju Kim, Ji-Eun Hong, Hyun-Jung Lee, Yeonyee~E Yoon, Jun-Bean Park, Seung-Pyo Lee, et~al. 2022.
\newblock \href {https://www.nature.com/articles/s41598-022-25467-w} {{Differential Diagnosis of Common Etiologies of Left Ventricular Hypertrophy Using a Hybrid CNN-LSTM Model}}.
\newblock \emph{Scientific Reports}, 12(1):20998.

\bibitem[{Jiang et~al.(2025)Jiang, Black, Geng, Park, Ng, and Chen}]{jiang2025medagentbench}
Yixing Jiang, Kameron~C Black, Gloria Geng, Danny Park, Andrew~Y Ng, and Jonathan~H Chen. 2025.
\newblock Medagentbench: Dataset for benchmarking llms as agents in medical applications.
\newblock \emph{arXiv preprint arXiv:2501.14654}.

\bibitem[{Johri et~al.(2025)Johri, Jeong, Tran, Schlessinger, Wongvibulsin, Barnes, Zhou, Cai, Van~Allen, Kim et~al.}]{johri2025craftmd}
Shreya Johri, Jaehwan Jeong, Benjamin~A Tran, Daniel~I Schlessinger, Shannon Wongvibulsin, Leandra~A Barnes, Hong-Yu Zhou, Zhuo~Ran Cai, Eliezer~M Van~Allen, David Kim, et~al. 2025.
\newblock \href {https://www.nature.com/articles/s41591-024-03328-5} {An evaluation framework for clinical use of large language models in patient interaction tasks}.
\newblock \emph{Nature Medicine}, pages 1--10.

\bibitem[{Johri et~al.(2024)Johri, Jeong, Tran, Schlessinger, Wongvibulsin, Cai, Daneshjou, and Rajpurkar}]{johri2024craft}
Shreya Johri, Jaehwan Jeong, Benjamin~A. Tran, Daniel~I Schlessinger, Shannon Wongvibulsin, Zhuo~Ran Cai, Roxana Daneshjou, and Pranav Rajpurkar. 2024.
\newblock \href {https://openreview.net/forum?id=Bk2nbTDtm8} {{CRAFT}-{MD}: A conversational evaluation framework for comprehensive assessment of clinical {LLM}s}.
\newblock In \emph{AAAI 2024 Spring Symposium on Clinical Foundation Models}.

\bibitem[{Kavanagh et~al.(2024)Kavanagh, Van Der~Tuijn, and Bain}]{2024PrinciplesOD}
Sallianne Kavanagh, Katie Van Der~Tuijn, and Amie Bain. 2024.
\newblock \href {https://doi.org/10.1211/PJ.2024.1.228037} {Principles of diagnostic reasoning}.
\newblock \emph{Pharmaceutical Journal}, 312(7982).

\bibitem[{Kim et~al.(2024)Kim, Park, Jeong, Chan, Xu, McDuff, Lee, Ghassemi, Breazeal, and Park}]{kim2024mdagents}
Yubin Kim, Chanwoo Park, Hyewon Jeong, Yik~Siu Chan, Xuhai~"Orson" Xu, Daniel McDuff, Hyeonhoon Lee, Marzyeh Ghassemi, Cynthia Breazeal, and Hae Park. 2024.
\newblock \href {https://proceedings.neurips.cc/paper_files/paper/2024/file/90d1fc07f46e31387978b88e7e057a31-Paper-Conference.pdf} {Mdagents: An adaptive collaboration of llms for medical decision-making}.
\newblock In \emph{Advances in Neural Information Processing Systems}, volume~37, pages 79410--79452. Curran Associates, Inc.

\bibitem[{Li et~al.(2024{\natexlab{a}})Li, Wang, Zhang, Li, Lai, Kang, Ma, and Liu}]{li2024agent}
Junkai Li, Siyu Wang, Meng Zhang, Weitao Li, Yunghwei Lai, Xinhui Kang, Weizhi Ma, and Yang Liu. 2024{\natexlab{a}}.
\newblock \href {https://arxiv.org/pdf/2405.02957} {Agent hospital: A simulacrum of hospital with evolvable medical agents}.
\newblock \emph{arXiv preprint arXiv:2405.02957}.

\bibitem[{Li et~al.(2024{\natexlab{b}})Li, Balachandran, Feng, Ilgen, Pierson, Koh, and Tsvetkov}]{li2024mediq}
Shuyue~Stella Li, Vidhisha Balachandran, Shangbin Feng, Jonathan~S. Ilgen, Emma Pierson, Pang~Wei Koh, and Yulia Tsvetkov. 2024{\natexlab{b}}.
\newblock \href {https://openreview.net/forum?id=W4pIBQ7bAI} {Mediq: Question-asking {LLM}s and a benchmark for reliable interactive clinical reasoning}.
\newblock In \emph{The Thirty-eighth Annual Conference on Neural Information Processing Systems}.

\bibitem[{Liu et~al.(2020)Liu, Jain, Eng, Way, Lee, Bui, Kanada, de~Oliveira~Marinho, Gallegos, Gabriele et~al.}]{liu2020deep}
Yuan Liu, Ayush Jain, Clara Eng, David~H Way, Kang Lee, Peggy Bui, Kimberly Kanada, Guilherme de~Oliveira~Marinho, Jessica Gallegos, Sara Gabriele, et~al. 2020.
\newblock \href {https://www.nature.com/articles/s41591-020-0842-3} {{A Deep Learning System for Differential Diagnosis of Skin Diseases}}.
\newblock \emph{Nature medicine}, 26(6):900--908.

\bibitem[{Maharjan et~al.(2024)Maharjan, Garikipati, Singh, Cyrus, Sharma, Ciobanu, Barnes, Thapa, Mao, and Das}]{maharjan2024openmedlm}
Jenish Maharjan, Anurag Garikipati, Navan~Preet Singh, Leo Cyrus, Mayank Sharma, Madalina Ciobanu, Gina Barnes, Rahul Thapa, Qingqing Mao, and Ritankar Das. 2024.
\newblock Openmedlm: prompt engineering can out-perform fine-tuning in medical question-answering with open-source large language models.
\newblock \emph{Scientific Reports}, 14(1):14156.

\bibitem[{McDuff et~al.(2023)McDuff, Schaekermann, Tu, Palepu, Wang, Garrison, Singhal, Sharma, Azizi, Kulkarni et~al.}]{mcduff2023towards}
Daniel McDuff, Mike Schaekermann, Tao Tu, Anil Palepu, Amy Wang, Jake Garrison, Karan Singhal, Yash Sharma, Shekoofeh Azizi, Kavita Kulkarni, et~al. 2023.
\newblock \href {https://arxiv.org/pdf/2312.00164} {{Towards Accurate Differential Diagnosis with Large Language Models}}.
\newblock \emph{arXiv preprint arXiv:2312.00164}.

\bibitem[{Mizuta et~al.(2024)Mizuta, Hirosawa, Harada, and Shimizu}]{mizuta2024diagnosis}
Kazuya Mizuta, Takanobu Hirosawa, Yukinori Harada, and Taro Shimizu. 2024.
\newblock \href {https://www.degruyter.com/document/doi/10.1515/dx-2024-0027/html} {Can chatgpt-4 evaluate whether a differential diagnosis list contains the correct diagnosis as accurately as a physician?}
\newblock \emph{Diagnosis}, (0).

\bibitem[{Nachane et~al.(2024)Nachane, Gramopadhye, Chanda, Ramakrishnan, Jadhav, Nandwani, Raghu, and Joshi}]{nachane2024few}
Saeel~Sandeep Nachane, Ojas Gramopadhye, Prateek Chanda, Ganesh Ramakrishnan, Kshitij~Sharad Jadhav, Yatin Nandwani, Dinesh Raghu, and Sachindra Joshi. 2024.
\newblock \href {https://doi.org/10.18653/v1/2024.findings-emnlp.31} {Few shot chain-of-thought driven reasoning to prompt {LLM}s for open-ended medical question answering}.
\newblock In \emph{Findings of the Association for Computational Linguistics: EMNLP 2024}, pages 542--573, Miami, Florida, USA. Association for Computational Linguistics.

\bibitem[{Nori et~al.(2023{\natexlab{a}})Nori, King, McKinney, Carignan, and Horvitz}]{nori2023capabilities}
Harsha Nori, Nicholas King, Scott~Mayer McKinney, Dean Carignan, and Eric Horvitz. 2023{\natexlab{a}}.
\newblock Capabilities of gpt-4 on medical challenge problems.
\newblock \emph{arXiv preprint arXiv:2303.13375}.

\bibitem[{Nori et~al.(2023{\natexlab{b}})Nori, Lee, Zhang, Carignan, Edgar, Fusi, King, Larson, Li, Liu et~al.}]{nori2023can}
Harsha Nori, Yin~Tat Lee, Sheng Zhang, Dean Carignan, Richard Edgar, Nicolo Fusi, Nicholas King, Jonathan Larson, Yuanzhi Li, Weishung Liu, et~al. 2023{\natexlab{b}}.
\newblock \href {https://arxiv.org/pdf/2311.16452} {Can generalist foundation models outcompete special-purpose tuning? case study in medicine}.
\newblock \emph{arXiv preprint arXiv:2311.16452}.

\bibitem[{Qiu et~al.(2024)Qiu, Wu, Zhang, Lin, Wang, Zhang, Wang, and Xie}]{qiu2024towards}
Pengcheng Qiu, Chaoyi Wu, Xiaoman Zhang, Weixiong Lin, Haicheng Wang, Ya~Zhang, Yanfeng Wang, and Weidi Xie. 2024.
\newblock \href {https://www.nature.com/articles/s41467-024-52417-z} {Towards building multilingual language model for medicine}.
\newblock \emph{Nature Communications}, 15(1):8384.

\bibitem[{Reese et~al.(2024)Reese, Danis, Caufield, Groza, Casiraghi, Valentini, Mungall, and Robinson}]{reese2024limitations}
Justin~T Reese, Daniel Danis, J~Harry Caufield, Tudor Groza, Elena Casiraghi, Giorgio Valentini, Christopher~J Mungall, and Peter~N Robinson. 2024.
\newblock \href {https://pmc.ncbi.nlm.nih.gov/articles/PMC10370243/} {On the limitations of large language models in clinical diagnosis}.
\newblock \emph{medRxiv}, pages 2023--07.

\bibitem[{Rhoads et~al.(2017)Rhoads, Penick et~al.}]{rhoads2017formulating}
Jacqueline Rhoads, Julie~C Penick, et~al. 2017.
\newblock \emph{Formulating a Differential Diagnosis for the Advanced Practice Provider}.
\newblock Springer Publishing Company.

\bibitem[{Rose et~al.(2023)Rose, Himakunthala, Ouyang, He, Mei, Lu, Saxon, Sonar, Mirza, and Wang}]{rose2023visual}
Daniel Rose, Vaishnavi Himakunthala, Andy Ouyang, Ryan He, Alex Mei, Yujie Lu, Michael Saxon, Chinmay Sonar, Diba Mirza, and William~Yang Wang. 2023.
\newblock \href {https://arxiv.org/pdf/2305.02317} {Visual chain of thought: bridging logical gaps with multimodal infillings}.
\newblock \emph{arXiv preprint arXiv:2305.02317}.

\bibitem[{Rutledge(2024)}]{rutledge2024diagnostic}
Geoffrey~W Rutledge. 2024.
\newblock \href {https://onlinelibrary.wiley.com/doi/full/10.1002/lrh2.10438} {{Diagnostic Accuracy of GPT-4 on Common Clinical Scenarios and Challenging Cases}}.
\newblock \emph{Learning Health Systems}, 8(3):e10438.

\bibitem[{Saab et~al.(2024)Saab, Tu, Weng, Tanno, Stutz, Wulczyn, Zhang, Strother, Park, Vedadi et~al.}]{saab2024capabilities}
Khaled Saab, Tao Tu, Wei-Hung Weng, Ryutaro Tanno, David Stutz, Ellery Wulczyn, Fan Zhang, Tim Strother, Chunjong Park, Elahe Vedadi, et~al. 2024.
\newblock Capabilities of gemini models in medicine.
\newblock \emph{arXiv preprint arXiv:2404.18416}.

\bibitem[{Savage et~al.(2024)Savage, Nayak, Gallo, Rangan, and Chen}]{savage2024diagnostic}
Thomas Savage, Ashwin Nayak, Robert Gallo, Ekanath Rangan, and Jonathan~H Chen. 2024.
\newblock \href {https://www.nature.com/articles/s41746-024-01010-1} {Diagnostic reasoning prompts reveal the potential for large language model interpretability in medicine}.
\newblock \emph{NPJ Digital Medicine}, 7(1):20.

\bibitem[{Schmidgall et~al.(2024)Schmidgall, Ziaei, Harris, Reis, Jopling, and Moor}]{schmidgall2024agentclinic}
Samuel Schmidgall, Rojin Ziaei, Carl Harris, Eduardo Reis, Jeffrey Jopling, and Michael Moor. 2024.
\newblock \href {https://arxiv.org/pdf/2405.07960} {Agentclinic: a multimodal agent benchmark to evaluate ai in simulated clinical environments}.
\newblock \emph{arXiv preprint arXiv:2405.07960}.

\bibitem[{Shikino et~al.(2024)Shikino, Shimizu, Otsuka, Tago, Takahashi, Watari, Sasaki, Iizuka, Tamura, Nakashima et~al.}]{shikino2024evaluation}
Kiyoshi Shikino, Taro Shimizu, Yuki Otsuka, Masaki Tago, Hiromizu Takahashi, Takashi Watari, Yosuke Sasaki, Gemmei Iizuka, Hiroki Tamura, Koichi Nakashima, et~al. 2024.
\newblock \href {https://mededu.jmir.org/2024/1/e58758} {{Evaluation of ChatGPT-Generated Differential Diagnosis for Common Diseases With Atypical Presentation: Descriptive Research}}.
\newblock \emph{JMIR Medical Education}, 10:e58758.

\bibitem[{Sun et~al.(2024)Sun, Luo, Liu, and Huang}]{sun2024conversational}
Zhoujian Sun, Cheng Luo, Ziyi Liu, and Zhengxing Huang. 2024.
\newblock \href {https://arxiv.org/pdf/2404.04292} {Conversational disease diagnosis via external planner-controlled large language models}.
\newblock \emph{arXiv preprint arXiv:2404.04292}.

\bibitem[{Tang et~al.(2024)Tang, Zou, Zhang, Li, Zhao, Zhang, Cohan, and Gerstein}]{tang-etal-2024-medagents}
Xiangru Tang, Anni Zou, Zhuosheng Zhang, Ziming Li, Yilun Zhao, Xingyao Zhang, Arman Cohan, and Mark Gerstein. 2024.
\newblock \href {https://doi.org/10.18653/v1/2024.findings-acl.33} {{M}ed{A}gents: Large language models as collaborators for zero-shot medical reasoning}.
\newblock In \emph{Findings of the Association for Computational Linguistics: ACL 2024}, pages 599--621, Bangkok, Thailand. Association for Computational Linguistics.

\bibitem[{Ten~Berg et~al.(2024)Ten~Berg, van Bakel, van~de Wouw, Jie, Schipper, Jansen, O’Connor, van Ginneken, and Kurstjens}]{ten2024chatgpt}
Hidde Ten~Berg, Bram van Bakel, Lieke van~de Wouw, Kim~E Jie, Anoeska Schipper, Henry Jansen, Rory~D O’Connor, Bram van Ginneken, and Steef Kurstjens. 2024.
\newblock \href {https://doi.org/10.1016/j.annemergmed.2023.08.003} {{ChatGPT and Generating a Differential Diagnosis Early in an Emergency Department Presentation}}.
\newblock \emph{Annals of Emergency Medicine}, 83(1):83--86.

\bibitem[{Tu et~al.(2024)Tu, Palepu, Schaekermann, Saab, Freyberg, Tanno, Wang, Li, Amin, Tomasev et~al.}]{tu2024towards}
Tao Tu, Anil Palepu, Mike Schaekermann, Khaled Saab, Jan Freyberg, Ryutaro Tanno, Amy Wang, Brenna Li, Mohamed Amin, Nenad Tomasev, et~al. 2024.
\newblock \href {https://arxiv.org/pdf/2401.05654} {Towards conversational diagnostic ai}.
\newblock \emph{arXiv preprint arXiv:2401.05654}.

\bibitem[{Wang et~al.(2024)Wang, Zhao, Qiang, Xi, Qin, and Liu}]{wang2024beyond}
Haochun Wang, Sendong Zhao, Zewen Qiang, Nuwa Xi, Bing Qin, and Ting Liu. 2024.
\newblock \href {https://arxiv.org/pdf/2401.16107} {Beyond direct diagnosis: Llm-based multi-specialist agent consultation for automatic diagnosis}.
\newblock \emph{arXiv preprint arXiv:2401.16107}.

\bibitem[{Wei et~al.(2022)Wei, Wang, Schuurmans, Bosma, Xia, Chi, Le, Zhou et~al.}]{wei2022chain}
Jason Wei, Xuezhi Wang, Dale Schuurmans, Maarten Bosma, Fei Xia, Ed~Chi, Quoc~V Le, Denny Zhou, et~al. 2022.
\newblock \href {https://proceedings.neurips.cc/paper_files/paper/2022/file/9d5609613524ecf4f15af0f7b31abca4-Paper-Conference.pdf} {Chain-of-thought prompting elicits reasoning in large language models}.
\newblock \emph{Advances in neural information processing systems}, 35:24824--24837.

\bibitem[{Winter et~al.(2024)Winter, Figueroa~Rosero, Loeser, Gers, Figueroa~Rosero, and Krestel}]{winter2024ddxgym}
Benjamin Winter, Alexei~Gustavo Figueroa~Rosero, Alexander Loeser, Felix~Alexander Gers, Nancy~Katerina Figueroa~Rosero, and Ralf Krestel. 2024.
\newblock \href {https://aclanthology.org/2024.lrec-main.396/} {{DD}x{G}ym: Online transformer policies in a knowledge graph based natural language environment}.
\newblock In \emph{Proceedings of the 2024 Joint International Conference on Computational Linguistics, Language Resources and Evaluation (LREC-COLING 2024)}, pages 4438--4448, Torino, Italia. ELRA and ICCL.

\bibitem[{Wu et~al.(2023)Wu, Chen, and Chen}]{wu2023large}
Cheng-Kuang Wu, Wei-Lin Chen, and Hsin-Hsi Chen. 2023.
\newblock \href {https://openreview.net/forum?id=N0lQfjeNWOE} {Large language models perform diagnostic reasoning}.

\bibitem[{Wu et~al.(2024)Wu, Tam, Lin, Chen, and yi~Lee}]{wu2024streambench}
Cheng-Kuang Wu, Zhi~Rui Tam, Chieh-Yen Lin, Yun-Nung Chen, and Hung yi~Lee. 2024.
\newblock \href {https://openreview.net/forum?id=8hUUy3hoS8} {Streambench: Towards benchmarking continuous improvement of language agents}.
\newblock In \emph{The Thirty-eight Conference on Neural Information Processing Systems Datasets and Benchmarks Track}.

\bibitem[{Xiao et~al.(2024)Xiao, Liu, Zhang, Muennighoff, Lian, and Nie}]{xiao-etal-2024-bge}
Shitao Xiao, Zheng Liu, Peitian Zhang, Niklas Muennighoff, Defu Lian, and Jian-Yun Nie. 2024.
\newblock \href {https://doi.org/10.1145/3626772.3657878} {C-pack: Packed resources for general chinese embeddings}.
\newblock In \emph{Proceedings of the 47th International ACM SIGIR Conference on Research and Development in Information Retrieval}, SIGIR '24, page 641–649, New York, NY, USA. Association for Computing Machinery.

\bibitem[{Yao et~al.(2023)Yao, Zhao, Yu, Du, Shafran, Narasimhan, and Cao}]{yao2023react}
Shunyu Yao, Jeffrey Zhao, Dian Yu, Nan Du, Izhak Shafran, Karthik~R Narasimhan, and Yuan Cao. 2023.
\newblock \href {https://openreview.net/forum?id=WE_vluYUL-X} {React: Synergizing reasoning and acting in language models}.
\newblock In \emph{The Eleventh International Conference on Learning Representations}.

\bibitem[{Zelin et~al.(2024)Zelin, Chung, Jeanne, Zhang, and Weng}]{zelin2024rare}
Charlotte Zelin, Wendy~K Chung, Mederic Jeanne, Gongbo Zhang, and Chunhua Weng. 2024.
\newblock \href {https://doi.org/10.1016/j.jbi.2024.104702} {Rare disease diagnosis using knowledge guided retrieval augmentation for chatgpt}.
\newblock \emph{Journal of Biomedical Informatics}, 157:104702.

\bibitem[{Zhang et~al.(2024)Zhang, Zeng, Hua, Ding, Chen, Ma, Li, Cui, Qi, Zhu, Lv, Jinfang, Liu, and Zhou}]{zhang2024ultramedical}
Kaiyan Zhang, Sihang Zeng, Ermo Hua, Ning Ding, Zhang-Ren Chen, Zhiyuan Ma, Haoxin Li, Ganqu Cui, Biqing Qi, Xuekai Zhu, Xingtai Lv, Hu~Jinfang, Zhiyuan Liu, and Bowen Zhou. 2024.
\newblock \href {https://openreview.net/forum?id=pUcTrjRLOM} {Ultramedical: Building specialized generalists in biomedicine}.
\newblock In \emph{The Thirty-eight Conference on Neural Information Processing Systems Datasets and Benchmarks Track}.

\bibitem[{Zhou et~al.(2024)Zhou, Lin, Ding, Wang, Melton, Zou, and Zhang}]{zhou2024interpretable}
Shuang Zhou, Mingquan Lin, Sirui Ding, Jiashuo Wang, Genevieve~B Melton, James Zou, and Rui Zhang. 2024.
\newblock \href {https://arxiv.org/pdf/2407.07330} {Interpretable differential diagnosis with dual-inference large language models}.
\newblock \emph{arXiv preprint arXiv:2407.07330}.

\bibitem[{Zhu et~al.(2024)Zhu, Ren, Wang, Zheng, Xie, Feng, Zhu, Li, Ma, and Pan}]{zhu2024emerge}
Yinghao Zhu, Changyu Ren, Zixiang Wang, Xiaochen Zheng, Shiyun Xie, Junlan Feng, Xi~Zhu, Zhoujun Li, Liantao Ma, and Chengwei Pan. 2024.
\newblock \href {https://arxiv.org/pdf/2406.00036} {Emerge: Integrating rag for improved multimodal ehr predictive modeling}.
\newblock \emph{arXiv preprint arXiv:2406.00036}.

\bibitem[{Zuo et~al.(2024)Zuo, Jiang, Mo, and Lio}]{zuo2024kg4diagnosis}
Kaiwen Zuo, Yirui Jiang, Fan Mo, and Pietro Lio. 2024.
\newblock \href {https://arxiv.org/pdf/2412.16833} {Kg4diagnosis: A hierarchical multi-agent llm framework with knowledge graph enhancement for medical diagnosis}.
\newblock \emph{arXiv preprint arXiv:2412.16833}.

\end{thebibliography}

\appendix
\clearpage
\onecolumn

\section{Details of DDx Benchmark}
\label{app:ddx_benchmark_details}

\setlength{\tabcolsep}{10.7pt}
\begin{table*}[h]
    \centering
    \footnotesize
\begin{tabular}{lccccc}
\toprule
\textbf{Dataset}   & \textbf{Domain}  & \textbf{\# Cases}    & \textbf{\# Diseases} & \textbf{Synthetic}   & \textbf{License\dag}    \\
\midrule
DDxPlus~\citep{fansi2022ddxplus}   & \textit{respiratory} & 1.3M    & 49           & \checkmark & CC-BY      \\
iCraft-MD~\citep{li2024mediq}   & \textit{skin}    & 140    & 394         &   \checkmark       & MIT        \\
RareBench~\citep{chen2024rarebench} & \textit{rare}    & 2,185    & 421         &     $\times$ 
& Apache-2.0 \\ 
\bottomrule
\end{tabular}
\caption{Overview of the selected sources for constructing DDx benchmark. We consider three domains (i.e., disease categories) (\textit{respiratory}, \textit{skin}, \textit{rare}) with different sizes of diagnosis options. All selected sources are applicable for \textit{commercial} usage. \dag License: Creative Commons Attribution International License (CC-BY). }
    \label{tab:datasets_overview}
\end{table*}

    \paragraph{Datasets.}
    To address the limitation of existing work, which often evaluates on a \textit{single} dataset and diagnoses in a \textit{single} turn, we construct a comprehensive DDx benchmark sourced from three datasets: DDxPlus, iCraft-MD, and RareBench, covering \textit{respiratory}, \textit{skin}, and \textit{rare} diseases, respectively. The statistics of each dataset are presented in~\autoref{tab:datasets_overview}. DDxPlus~\citep{fansi2022ddxplus} is a large-scale synthetic dataset, spanning 1.3 million patient cases across 49 respiratory-related pathologies, focusing on conditions where the chief complaint is related to cough, sore through, or breathing issues. As one of the largest structured DDx datasets, it provides both ground-truth diagnoses \textit{and} ground truth ranked differential diagnosis lists, enabling effective few shot examples as well as a direct evaluation of predicting and refining DDx rankings. iCraft-MD (or interactive Craft-MD)~\citep{li2024mediq} adapts static dermatological clinical vignettes from the original Craft-MD dataset~\citep{johri2024craft, johri2025craftmd} into an interactive setting. It consists of 140 dermatology cases, with 100 sourced from an online medical question bank and 40 designed by expert clinicians. RareBench~\citep{chen2024rarebench} further expands the diagnostic landscape by extending DDxPlus to include 421 rare diseases. We specifically select three regional subsets from Rarebench -- RAMEDIS (Europe), MME (Canada), and PUMCH (China) -- to ensure diversity in rare disease regional representation. Each of these datasets includes patient profiles with two core components: (1) symptom/antecedent data and (2) ground-truth pathology/disease.

    \paragraph{Benchmark Compilation.}
    
    To enable a consistent evaluation across datasets, we normalize each dataset into a structured format, each dataset is converted to include: (i) Optional initial information of the patient (e.g., age, sex, chief complaint); (ii) full patient profile (complete list of symptoms and medical history); (iii) full set of possible diseases for differential diagnosis. 
For DDxPlus, we inherit the format from StreamBench~\citep{wu2024streambench}. In iCraft-MD, the initial information of the patient is provided as initial case details, whereas in Rarebench no initial patient information is available. We process iCraft-MD and RareBench to extract the full patient profile. One of the major challenges in iCraft-MD and Rarebench is the lack of predefined differential diagnosis options and the presence of redundant disease names. To address this, we iterate through all patient records and employ GPT-4o to generate a unique, non-redundant disease set for each patient case. As a result, we obtain 394 unique dermatological conditions for iCraft-MD and 102 rare diseases for the RareBench subset. This refinement step ensures that each patient’s diagnostic process operates within a well-defined differential diagnosis structure, reducing ambiguity and improving evaluation reliability.
\newpage
\begin{example}{\textbf{DDxPlus}}
\textbf{Initial Patient Profile:}\\
Age: 39\\
Sex: M\\
Chief Complaint: nasal congestion\\
\textbf{Complete Patient Profile: }\\
Sex: Male, Age: 39\\
- I am currently being treated or have recently been treated with an oral antibiotic for an ear infection.\\
- I have pain somewhere related to my reason for consulting.\\
- I have a fever (either felt or measured with a thermometer).\\
- I have nasal congestion or a clear runny nose.\\
- My vaccinations are up to date.\\
- On a scale of 0-10, the pain intensity is 6\\
- On a scale of 0-10, the pain's location precision is 8\\
- On a scale of 0-10, the pace at which the pain appear is 2\\
- The pain is:\\
* sensitive\\
* sharp\\
- The pain locations are:\\
* ear(R)\\
- The pain radiates to these locations:\\
* nowhere\\
\textbf{Ground Truth Pathology: } Acute otitis media \\
\textbf{Ground Truth DDx:} \\
1. Acute otitis media\\
2. URTI\\
3. Chagas
\end{example}
\begin{example}{\textbf{iCraft-MD}}
\textbf{Initial Patient Profile: }\\
Age: 61 years\\
Sex: male\\
Chief Complaint: A 61-year-old man presents with a 7-month history of lesions on his hands and arms\\
\textbf{Complete Patient Profile: } \\
- A 61-year-old man presents with a 7-month history of lesions on his hands and arms\\
- His medical history includes depression, hypertension, and hyperlipidemia\\
- He has no personal or family history of skin problems\\
- His skin lesions are not painful or itchy, and he is not bothered by their appearance\\
- He has not tried any treatments for the lesions\\
- Physical examination reveals a number of pink, annular plaques with smooth raised borders on the patient’s dorsal forearms and hands\\
- On close inspection, small discrete papules are seen within the plaques.\\
\textbf{Ground Truth Pathology: } Localized granuloma annulare
\end{example}

\begin{example}{\textbf{RareBench}}
\textbf{Initial Patient Profile: } \\
N/A\\
\textbf{Complete Patient Profile: }
- Hematuria \\
- Slurred speech \\
- Abnormality of the liver \\
- Dysphagia \\
- Drooling \\
- Abnormal caudate nucleus morphology \\
- Hand tremor \\
- Poor appetite \\
- Decreased circulating ceruloplasmin concentration \\
- Increased urinary copper concentration \\
- Kayser-Fleischer ring \\
\textbf{Ground Truth Pathology: } Wilson disease
\end{example}
\newpage
\section{Prompt Design}
We present the prompt design for history taking simulator, knowledge retrieval agent, diagnosis strategy agent, and DDxDriver in this section.

\begin{prompt}{\textbf{History Taking Simulator: Doctor}}
\textbf{System Prompt:}\\
\verb|<SPECIALIST_PREFACE>|\\
Your job is to take medical history from a patient by asking them specific questions to determine their antecedents and symptoms, as well as narrow down the possible diseases they may be suffering from. \verb|[...]|\\

You may receive this additional information to guide your dialogue:\\
- Initial Patient Information: Information the patient has already self-reported, such as chief complaint, age, sex, etc.\\
- Dialogue History: The conversation you and the patient have had so far, formatted as \verb|'Doctor'| / \verb|'Patient'| turns.\\
- Suggested Conversation Goals: Specific topics or questions to try to cover in the dialogue. You may also ask questions outside of these conversation goals; do not limit yourself to these.\\

You may either start, end, or continue the conversation, as explained below:\\
\verb|[...]|\\

Response Instructions:\\
\verb|[...]|
\end{prompt}
\begin{prompt}{\textbf{History Taking Simulator: Patient}}
\textbf{System Prompt}
Act as a patient with the patient profile below engaging in a medical history taking with a doctor. \verb|[...]|\\

You may receive this additional information to guide your dialogue:\\
- Initial Patient Information: Information you as the patient have already self-reported to the doctor, such as chief complaint, age, sex, etc.\\
- Dialogue History: The conversation you and the patient have had so far, formatted as \verb|'Doctor'| / \verb|'Patient'| turns.\\

When asked for information which is explicitly present in your patient profile (including as synonyms), either respond:\\
    a.~ Positively ("Yes"...) if your patient profile explicitly indicates you have this antecedent/symptom\\
    b.~Negatively ("No"...) if your patient profile explicitly indicates that you do not have this antecedent/symptom\\

When asked for information which is not explicitly mentioned in your patient profile (including as synonyms), respond "I don't know." \\
\verb|[...]|\\

Response instructions:\\
\verb|[...]|
\end{prompt}

\begin{prompt}{\textbf{Knowledge Retrieval Agent}}
\textbf{Keywords Prompt:}\\
Your job is to assist in the creation of a differential diagnosis for a patient by searching for relevant information online. Given an input search from a user, break it up into a list of simplified keyword searches to find relevant medical information online. \\
Follow these steps:\\
\verb|[...]|\\
Format example:\\
Input search:\\
\verb|<INPUT_SEARCH>|\\
Keyword searches list:\\
\verb|[<KEYWORD_SEARCH_1>, <KEYWORD_SEARCH_2>]|\\

\textbf{Synthesis Prompt}\\
You are a helpful research assistant to a doctor creating a differential diagnosis of a patient. Concisely answer the doctor's input search by analyzing and summarizing the relevant medical content in the search results. \verb|[...]|\\

Inputs:\\
1. Doctor's Input Search: the search the doctor requested\\
- This search may contain multiple topics\\
2. Search results: the search results fetched\\
- You may only answer based on topics present in these search results\\
3. Diagnosis Options (optional): the possible diseases the patient may be suffering from\\
- If provided, use this exact terminology to refer to the diseases\\

Response Instructions:\\
\verb|[...]|
\end{prompt}

\begin{prompt}{\textbf{Diagnosis Strategy Agent}}
\textbf{System Prompt:}\\
\verb|<SPECIALIST_PREFACE>|\\
Given a patient's profile (a list of antecedents and symptoms), provide a ranked differential diagnosis of the \verb|<DDX_LENGTH>| most likely diseases. You may be provided a list of diagnosis options you can choose from. You must use this exact disease terminology when referring to the diseases. If you aren't provided the diagnosis options, consider all possible diseases. \\

Your ranked differential diagnosis should have the possible diseases ranked from most likely to least likely.\\

You will also be provided with:\\
1. Previous Ranked Differential Diagnoses: \verb|[...]|\\
2. Suggested Diagnosis Instructions (optional): \verb|[...]|\\
3. Previous Search Content (optional): \verb|[...]|\\
4. Patient profile: the known symptoms/antecedents of the patient\\
5: Patient examples (optional): \verb|[...]|\\

\verb|[...]|\\

Directly provide the ranked differential diagnosis of the \verb|<DDX_LENGTH>| most likely diseases for the patient in the following format (without additional text before or after), with one diagnosis per line (replace \verb|[DIAGNOSIS_X]| with the actual diagnosis name, and do not include the brackets themselves): \verb|[RANK_NUMBER]. [DIAGNOSIS]|. \verb|I.e.|: \\
1. \verb|[DIAGNOSIS_1]|\\
2. \verb|[DIAGNOSIS_2]|\\
...\\
Directly provide your response in the format specified, without additional text.
\end{prompt}

\begin{prompt}{\textbf{DDxDriver}}
\textbf{Fixed Iteration System Prompt:}\\
Your job is to facilitate the process of differential diagnosis of a patient by concisely prompting medical agents.\\  

You will be provided with: \\
1) Agent Descriptions. This includes:\\
a) Agent Function: A description of the function of medical agent.\\
b) Agent Prompt: A description of how to prompt the agent\\
2) Available Information: The available information you can extract from to prompt the agent. Do not invent new information. This may include: \\
    a) Patient Initial Information\\
    b) Patient Profile\\
    c) Dialogue History\\
    d) Previous RAG content\\
    - External information found about diseases the patient may be suffering from\\
    e) Previous Ranked Differential Diagnoses\\
    f) Diagnosis Options\\
    - These are the only diseases the patient may be suffering from.\\
    - You must use the exact terminology in this list when referring to the diseases\\

Follow these steps to create a prompt for the medical agent:\\
1. Analyze the description of the medical agent and its input prompt. Note whether its input prompt is optional or mandatory.\\
2. Review the current information you were provided. Determine how this information can help the agent.\\
- You should only prompt based on this current information.\\
3. Follow the agent's input prompt description and design a prompt for this agent.\\
4. Respond with your agent prompt, nothing else.
\end{prompt}

\newpage
\section{Additional Analysis}

\begin{figure*}[h]
    \centering
    \begin{subfigure}{0.48\textwidth}
    \includegraphics[trim={0cm 0cm 0cm 0cm },clip, width=\textwidth]{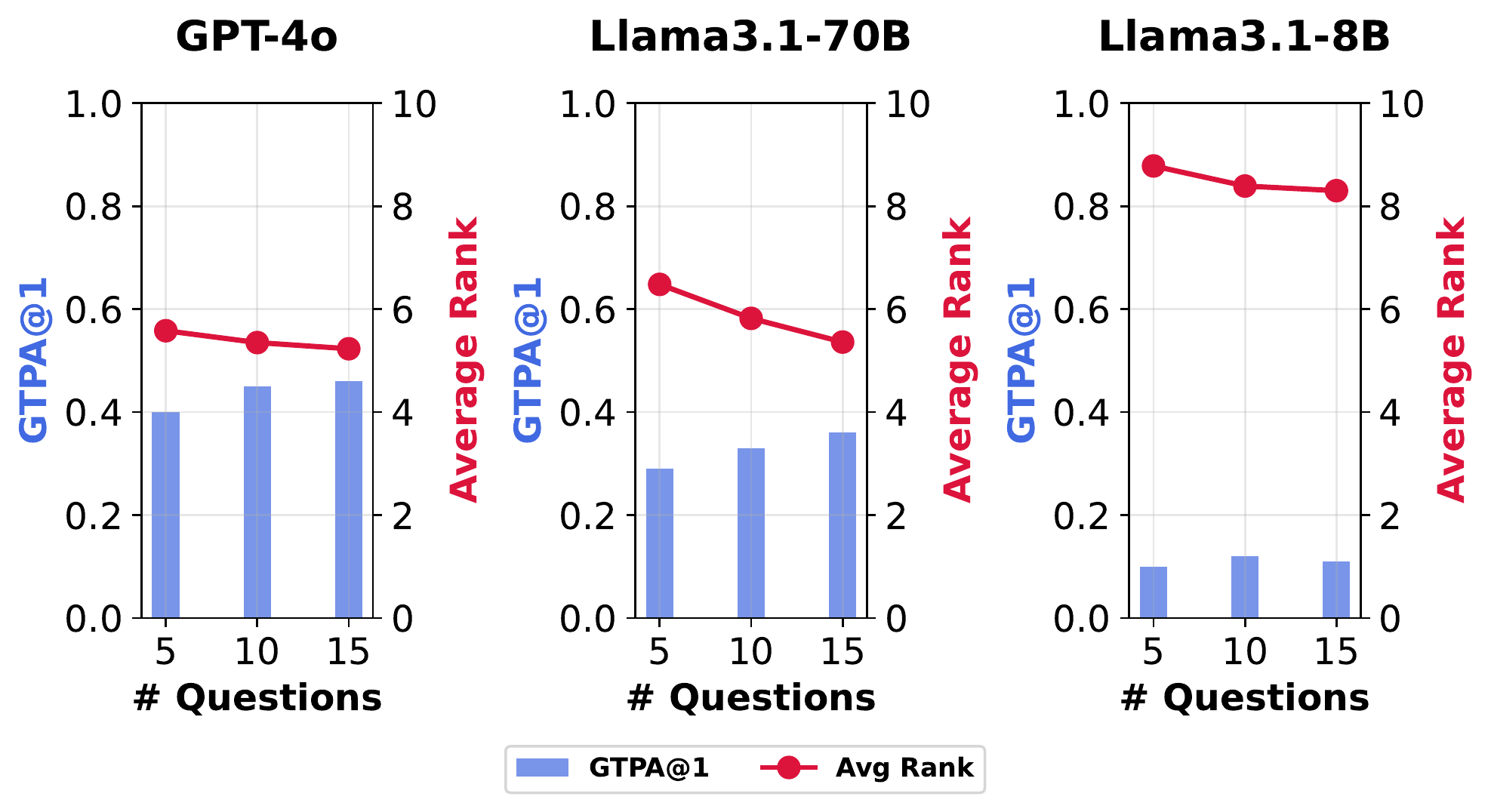}
    \caption{}
    \end{subfigure}
    \begin{subfigure}{0.48\textwidth}
    \includegraphics[trim={0cm 0cm 0cm 0cm}, clip, width=\textwidth]{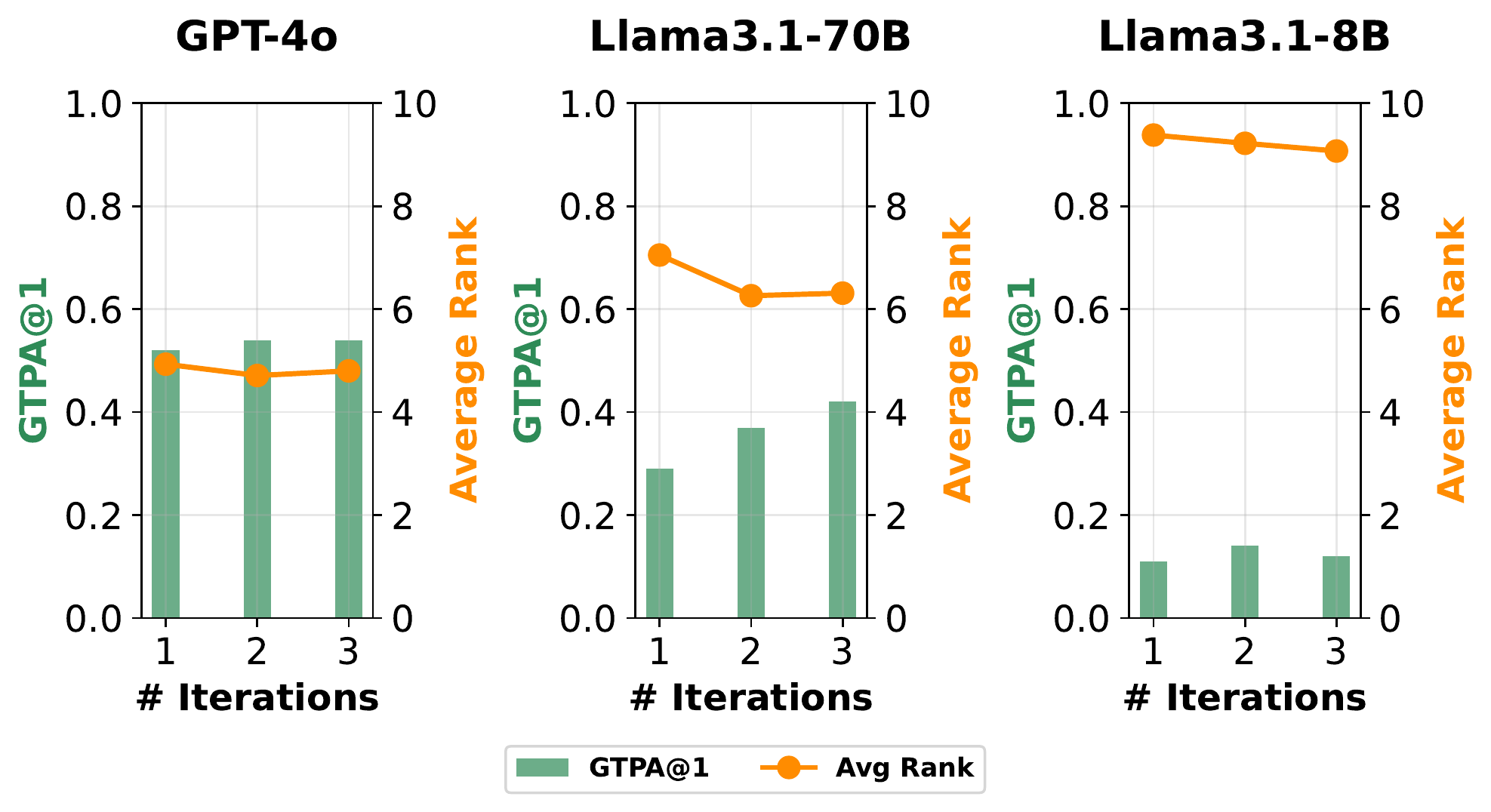}
    \caption{}
    \end{subfigure}
    \caption{Results of iCraft-MD~\citep{li2024mediq} compared between (a) History Taking Simulator, and (b) MEDDxAgent.}
    \label{fig:icraftmd_comparison}
\end{figure*}

\begin{figure*}[h]
    \centering
    \begin{subfigure}{0.48\textwidth}
    \includegraphics[trim={0cm 0cm 0cm 0cm },clip, width=\textwidth]{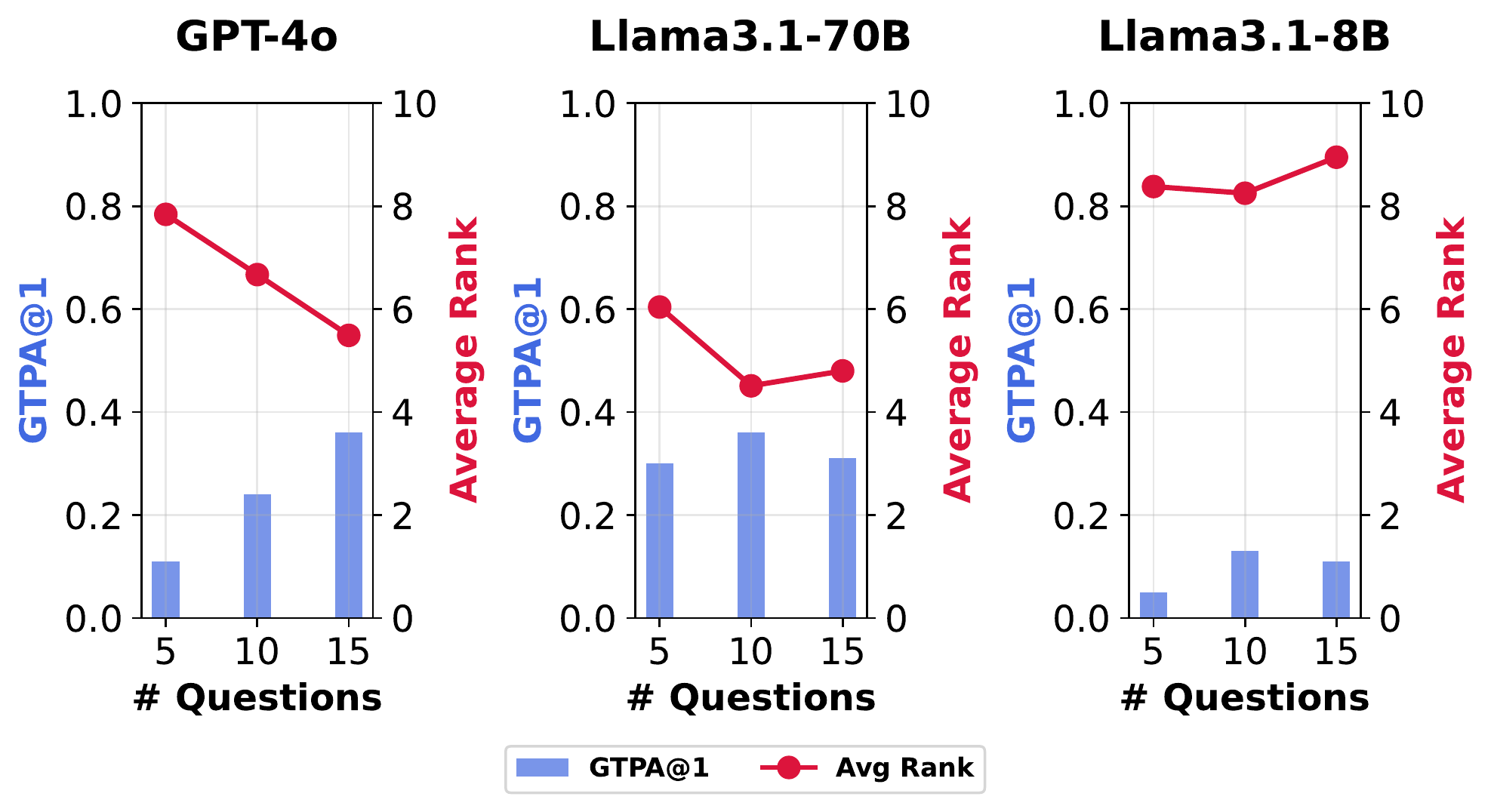}
    \caption{}
    \end{subfigure}
    \begin{subfigure}{0.48\textwidth}
    \includegraphics[trim={0cm 0cm 0cm 0cm}, clip, width=\textwidth]{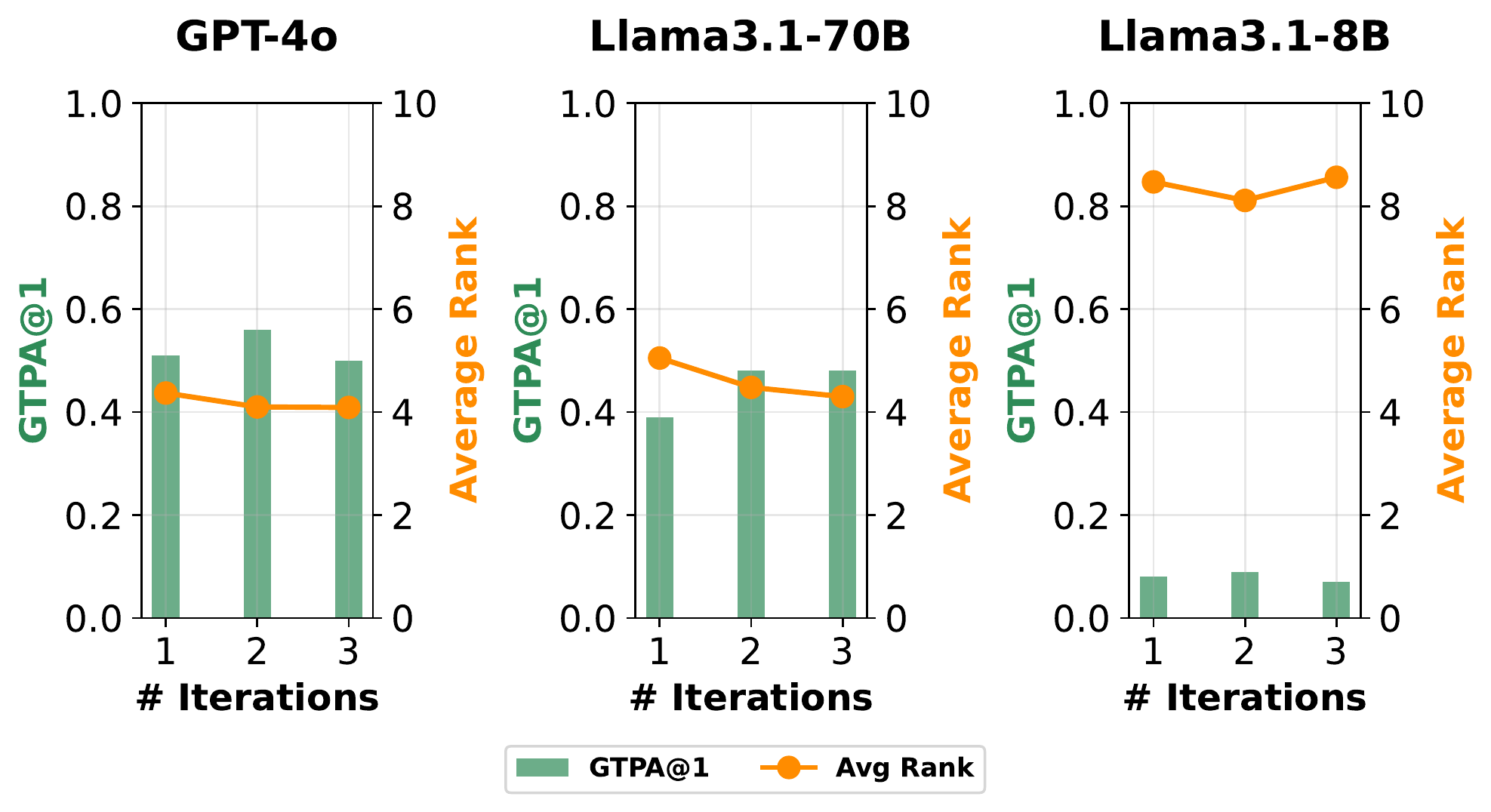}
    \caption{}
    \end{subfigure}
    \caption{Results of RareBench~\citep{chen2024rarebench} compared between (a) History Taking Simulator, and (b) MEDDxAgent.}
    \label{fig:rarebench_comparison}
\end{figure*}

Results of iCraft-MD (\autoref{fig:icraftmd_comparison}) and RareBench (\autoref{fig:rarebench_comparison}) compared between (a) History Taking Simulator, and (b) MEDDxAgent over the selection of max questions (5, 10, 15), and number of iterations (1, 2, 3).
\label{subsec:comparison_history_taking_iterative}

\newpage
\section{Additional Experiments}
\label{app:additional_experiments}
\subsection{History Taking Simulator}
\setlength{\tabcolsep}{11.6pt}
\begin{table*}[h]
    \centering
    \scriptsize{%
    \begin{tabular}{llcccccccccc}
    \toprule
     &  & \multicolumn{3}{c}{\textbf{DDxPlus}} & \multicolumn{3}{c}{\textbf{iCraft-MD}} & \multicolumn{3}{c}{\textbf{RareBench}} \\
    \cmidrule(lr){3-5} \cmidrule(lr){6-8} \cmidrule(lr){9-11}
    \textbf{Model} & \textbf{Metric} & \textbf{5} & \textbf{10} & \textbf{15} & \textbf{5} & \textbf{10} & \textbf{15} & \textbf{5} & \textbf{10} & \textbf{15} \\
    \midrule
    \multirow{4}{*}{\textbf{GPT-4o}}
    & \textbf{GTPA@1} & 0.45 & 0.59 & 0.69 & 0.40 & 0.45 & 0.46 & 0.11 & 0.24 & 0.36 \\
    & \textbf{GTPA@3} & 0.60 & 0.73 & 0.82 & 0.51 & 0.53 & 0.53 & 0.22 & 0.36 & 0.48 \\
    & \textbf{GTPA@5} & 0.72 & 0.83 & 0.88 & 0.57 & 0.57 & 0.60 & 0.35 & 0.47 & 0.59 \\
    & \textbf{Avg Rank} & 4.13 & 3.16 & 2.47 & 5.58 & 5.35 & 5.23 & 7.84 & 6.67 & 5.49 \\
    \midrule
    \multirow{4}{*}{\textbf{Llama3.1-70B}}
    & \textbf{GTPA@1} & 0.45 & 0.58 & 0.56 & 0.29 & 0.33 & 0.36 & 0.30 & 0.36 & 0.31 \\
    & \textbf{GTPA@3} & 0.65 & 0.77 & 0.76 & 0.39 & 0.49 & 0.53 & 0.42 & 0.57 & 0.59 \\
    & \textbf{GTPA@5} & 0.71 & 0.83 & 0.79 & 0.47 & 0.55 & 0.60 & 0.53 & 0.65 & 0.67 \\
    & \textbf{Avg Rank} & 4.15 & 3.12 & 3.50 & 6.48 & 5.82 & 5.36 & 6.04 & 4.51 & 4.80 \\
    \midrule
    \multirow{4}{*}{\textbf{UltraMedical-70B}}
    & \textbf{GTPA@1} & 0.45 & 0.52 & 0.50 & 0.23 & 0.27 & 0.22 & 0.40 & 0.43 & 0.44 \\
    & \textbf{GTPA@3} & 0.62 & 0.68 & 0.69 & 0.27 & 0.31 & 0.29 & 0.58 & 0.61 & 0.59 \\
    & \textbf{GTPA@5} & 0.70 & 0.73 & 0.73 & 0.29 & 0.34 & 0.31 & 0.61 & 0.67 & 0.67 \\
    & \textbf{Avg Rank} & 4.70 & 4.30 & 4.68 & 8.12 & 7.61 & 7.99 & 5.01 & 4.59 & 4.54 \\
    \midrule
    \multirow{5}{*}{\textbf{Llama3.1-8B}}
    & \textbf{GTPA@1} & 0.23 & 0.35 & 0.40 & 0.10 & 0.12 & 0.11 & 0.05 & 0.13 & 0.11 \\
    & \textbf{GTPA@3} & 0.37 & 0.49 & 0.53 & 0.20 & 0.23 & 0.23 & 0.21 & 0.21 & 0.18 \\
    & \textbf{GTPA@5} & 0.43 & 0.60 & 0.60 & 0.25 & 0.27 & 0.29 & 0.27 & 0.25 & 0.20 \\
    & \textbf{Avg Rank} & 6.85 & 5.46 & 5.44 & 8.78 & 8.39 & 8.3 & 8.38 & 8.25 & 8.95 \\
    \midrule
    \multirow{5}{*}{\textbf{UltraMedical3.1-8B}}
    & \textbf{GTPA@1} & 0.26 & 0.24 & 0.20 & 0.16 & 0.14 & 0.14 & 0.15 & 0.14 & 0.12 \\
    & \textbf{GTPA@3} & 0.36 & 0.35 & 0.32 & 0.22 & 0.20 & 0.21 & 0.23 & 0.25 & 0.21 \\
    & \textbf{GTPA@5} & 0.42 & 0.39 & 0.38 & 0.23 & 0.23 & 0.22 & 0.25 & 0.28 & 0.23 \\
    & \textbf{Avg Rank} & 7.00 & 7.86 & 7.73 & 8.60 & 8.78 & 8.72 & 8.44 & 8.29 & 8.46 \\
    \bottomrule
    \end{tabular}%
    }
    \caption{History taking simulator performance across different datasets (DDxPlus, iCraftMD, Rarebench) with max questions (5, 10, 15), aggregated over 100 patients.}
    \label{tab:performance_history_full}
\end{table*}
\newpage
\subsection{Knowledge Retrieval Agent}
\setlength{\tabcolsep}{15.2pt}
\begin{table*}[h]
    \centering
    \scriptsize{%
    \begin{tabular}{lllccc@{\hspace{10pt}}ccccc@{\hspace{10pt}}ccccc}
    \toprule
     & &  & \multicolumn{2}{c}{\textbf{DDxPlus}} & \multicolumn{2}{c}{\textbf{iCraft-MD}} & \multicolumn{2}{c}{\textbf{RareBench}} \\
    \cmidrule(l{.3em}r{.3em}){4-5} \cmidrule(l{.3em}r{.3em}){6-7} \cmidrule(l{.3em}r{.3em}){8-9}
     \textbf{Model}& \textbf{Source}& \textbf{Metric} & \textbf{RAG} & \textbf{Base} & \textbf{RAG} & \textbf{Base} & \textbf{RAG} & \textbf{Base} \\
    \midrule
    \multirow{8}{*}{\textbf{GPT-4o}} 
    & \multirow{4}{*}{\textbf{PubMed}} 
    & \textbf{GTPA@1} & 0.69 & 0.69 & 0.68 & 0.68 & 0.45 & 0.39 \\
    & & \textbf{GTPA@3} & 0.88 & 0.88 & 0.77 & 0.76 & 0.60 & 0.58 \\
    & & \textbf{GTPA@5} & 0.90 & 0.90 & 0.79 & 0.77 & 0.72 & 0.72 \\
    & & \textbf{Avg Rank} & 2.27 & 2.21 & 3.23 & 3.37 & 3.92 & 3.99 \\
    \cmidrule{2-9}
    & \multirow{4}{*}{\textbf{Wiki}} 
    & \textbf{GTPA@1} & 0.69 & 0.69 & 0.69 & 0.68 & 0.45 & 0.39 \\
    & & \textbf{GTPA@3} & 0.88 & 0.88 & 0.77 & 0.76 & 0.58 & 0.58 \\
    & & \textbf{GTPA@5} & 0.90 & 0.90 & 0.79 & 0.77 & 0.74 & 0.72 \\
    & & \textbf{Avg Rank} & 2.24 & 2.21 & 3.22 & 3.37 & 4.00 & 3.99 \\
    \midrule
    \multirow{8}{*}{\textbf{Llama3.1-70B}} 
    & \multirow{4}{*}{\textbf{PubMed}} 
    & \textbf{GTPA@1} & 0.56 & 0.54 & 0.44 & 0.40 & 0.38 & 0.39 \\
    & & \textbf{GTPA@3} & 0.77 & 0.74 & 0.56 & 0.56 & 0.62 & 0.59 \\
    & & \textbf{GTPA@5} & 0.79 & 0.78 & 0.63 & 0.64 & 0.75 & 0.77 \\
    & & \textbf{Avg Rank} & 3.42 & 3.53 & 4.72 & 4.87 & 3.96 & 4.05 \\
    \cmidrule{2-9}
    & \multirow{4}{*}{\textbf{Wiki}} 
    & \textbf{GTPA@1} & 0.49 & 0.54 & 0.44 & 0.40 & 0.39 & 0.39 \\
    & & \textbf{GTPA@3} & 0.74 & 0.74 & 0.59 & 0.56 & 0.59 & 0.59 \\
    & & \textbf{GTPA@5} & 0.77 & 0.78 & 0.66 & 0.64 & 0.75 & 0.77 \\
    & & \textbf{Avg Rank} & 3.60 & 3.53 & 4.71 & 4.87 & 4.09 & 4.05 \\
    \midrule
    \multirow{8}{*}{\textbf{UltraMedical-70B}} 
    & \multirow{4}{*}{\textbf{PubMed}} 
    & \textbf{GTPA@1} & 0.58 & 0.60 & 0.31 & 0.31 & 0.45 & 0.44 \\
    & & \textbf{GTPA@3} & 0.68 & 0.73 & 0.37 & 0.38 & 0.65 & 0.63 \\
    & & \textbf{GTPA@5} & 0.70 & 0.76 & 0.38 & 0.39 & 0.71 & 0.70 \\
    & & \textbf{Avg Rank} & 4.63 & 6.55 & 7.08 & 7.01 & 4.20 & 4.47 \\
    \cmidrule{2-9}
    & \multirow{4}{*}{\textbf{Wiki}} 
    & \textbf{GTPA@1} & 0.58 & 0.6 & 0.31 & 0.31 & 0.44 & 0.44 \\
    & & \textbf{GTPA@3} & 0.68 & 0.73 & 0.38 & 0.38 & 0.64 & 0.63 \\
    & & \textbf{GTPA@5} & 0.70 & 0.76 & 0.39 & 0.39 & 0.70 & 0.70 \\
    & & \textbf{Avg Rank} & 4.38 & 6.55 & 7.01 & 7.01 & 4.25 & 4.47 \\
    \midrule
    \multirow{8}{*}{\textbf{Llama3.1-8B}} 
    & \multirow{4}{*}{\textbf{PubMed}} 
    & \textbf{GTPA@1} & 0.42 & 0.48 & 0.29 & 0.27 & 0.35 & 0.33 \\
    & & \textbf{GTPA@3} & 0.58 & 0.63 & 0.38 & 0.37 & 0.55 & 0.54 \\
    & & \textbf{GTPA@5} & 0.67 & 0.69 & 0.44 & 0.43 & 0.59 & 0.57 \\
    & & \textbf{Avg Rank} & 4.50 & 5.25 & 6.93 & 7.02 & 5.33 & 5.45 \\
    \cmidrule{2-9}
    & \multirow{4}{*}{\textbf{Wiki}} 
    & \textbf{GTPA@1} & 0.43 & 0.48 & 0.29 & 0.27 & 0.36 & 0.33 \\
    & & \textbf{GTPA@3} & 0.58 & 0.63 & 0.38 & 0.37 & 0.52 & 0.54 \\
    & & \textbf{GTPA@5} & 0.67 & 0.69 & 0.44 & 0.43 & 0.67 & 0.57 \\
    & & \textbf{Avg Rank} & 4.56 & 5.25 & 6.93 & 7.02 & 4.80 & 5.45 \\
    \midrule
    \multirow{8}{*}{\textbf{UltraMedical3.1-8B}} 
    & \multirow{4}{*}{\textbf{PubMed}} 
    & \textbf{GTPA@1} & 0.27 & 0.33 & 0.19 & 0.27 & 0.21 & 0.22 \\
    & & \textbf{GTPA@3} & 0.39 & 0.48 & 0.24 & 0.37 & 0.46 & 0.35 \\
    & & \textbf{GTPA@5} & 0.46 & 0.51 & 0.26 & 0.43 & 0.48 & 0.42 \\
    & & \textbf{Avg Rank} & 6.81 & 6.88 & 8.38 & 7.02 & 6.23 & 7.02 \\
    \cmidrule{2-9}
    & \multirow{4}{*}{\textbf{Wiki}} 
    & \textbf{GTPA@1} & 0.25 & 0.33 & 0.18 & 0.18 & 0.27 & 0.22 \\
    & & \textbf{GTPA@3} & 0.38 & 0.48 & 0.23 & 0.23 & 0.44 & 0.35 \\
    & & \textbf{GTPA@5} & 0.45 & 0.51 & 0.25 & 0.25 & 0.51 & 0.42 \\
    & & \textbf{Avg Rank} & 6.90 & 6.88 & 8.42 & 8.45 & 6.37 & 7.02 \\
    \bottomrule
    \end{tabular}%
    }
    \caption{Knowledge retrieval agent performance with different datasets with varying sources (PubMed, Wikipedia) and methods, aggregated over 100 patients.}
    \label{tab:performance_rag_full}
\end{table*}
\newpage
\subsection{Diagnosis Strategy Agent}
\setlength{\tabcolsep}{2.5pt}
\begin{table*}[h]
    \centering
    \scriptsize
    \begin{tabular}{l l cccc cccc cccc}
    \toprule
    & & \multicolumn{4}{c}{\textbf{DDxPlus}} & \multicolumn{4}{c}{\textbf{iCraft-MD}} & \multicolumn{4}{c}{\textbf{RareBench}} \\
    \cmidrule(lr){3-6} \cmidrule(lr){7-10} \cmidrule(lr){11-14}
    \textbf{Model} & \textbf{Metric} & \textbf{None} & \textbf{Static} & \textbf{Dyn\_BAII} & \textbf{Dyn\_BERT} 
                              & \textbf{None} & \textbf{Static} & \textbf{Dyn\_BAII} & \textbf{Dyn\_BERT} 
                              & \textbf{None} & \textbf{Static} & \textbf{Dyn\_BAII} & \textbf{Dyn\_BERT} \\
                              
    \midrule
    \multicolumn{14}{l}{\textit{\textbf{Standard}}}\\
    \midrule
    \multirow{3}{*}{\textbf{GPT-4o}} 
    & \textbf{GTPA@1} & 0.69 & 0.74 & 0.96 & 0.96 & 0.68 & 0.64 & 0.62 & 0.67 & 0.46 & 0.52 & 0.79 & 0.78 \\
    & \textbf{GTPA@5} & 0.90 & 0.90 & 1.00 & 1.00 & 0.77 & 0.74 & 0.72 & 0.77 & 0.72 & 0.80 & 0.91 & 0.90 \\
    & \textbf{Avg Rank} & 2.21 & 2.20 & 1.06 & 1.06 & 3.37 & 3.64 & 3.85 & 3.31 & 3.99 & 3.58 & 2.03 & 2.19 \\
    \midrule

    \multirow{3}{*}{\textbf{Llama3.1-70B}} 
    & \textbf{GTPA@1} & 0.54 & 0.57 & 0.86 & 0.84 & 0.40 & 0.38 & 0.40 & 0.40 & 0.39 & 0.42 & 0.73 & 0.72 \\
    & \textbf{GTPA@5} & 0.78 & 0.80 & 0.95 & 0.94 & 0.64 & 0.63 & 0.63 & 0.62 & 0.77 & 0.71 & 0.87 & 0.87 \\
    & \textbf{Avg Rank} & 3.53 & 3.41 & 1.59 & 1.68 & 4.87 & 5.15 & 5.02 & 4.96 & 4.05 & 4.29 & 2.44 & 2.44 \\
     \midrule
    \multirow{3}{*}{\textbf{UltraMedical-70B}}  
    & \textbf{GTPA@1} & 0.58 & 0.60& 0.97 & 0.96 & 0.31 & 0.37 & 0.42 & 0.40& 0.44 & 0.47 & 0.74 & 0.71 \\
    & \textbf{GTPA@5} & 0.70 & 0.76 & 1.00& 1.00& 0.39 & 0.45 & 0.47 & 0.48 & 0.70& 0.62 & 0.83 & 0.80\\
    & \textbf{Avg Rank} & 4.18 & 6.55 & 1.03 & 1.04 & 7.01 & 6.29 & 6.14 & 6.15 & 4.47 & 4.92 & 2.74 & 2.96 \\
    \midrule
    \multirow{3}{*}{\textbf{Llama3.1-8B}}
    & \textbf{GTPA@1} & 0.45 & 0.48 & 0.97 & 0.97 & 0.27 & 0.25 & 0.21 & 0.22 & 0.33 & 0.39 & 0.71 & 0.70\\
    & \textbf{GTPA@5} & 0.68 & 0.69 & 1.00& 1.00& 0.43 & 0.44 & 0.42 & 0.40& 0.57 & 0.63 & 0.83 & 0.81 \\
    & \textbf{Avg Rank} & 9.00 & 5.25 & 1.03 & 1.04 & 7.02 & 6.78 & 6.93 & 7.32 & 5.45 & 4.76 & 2.80& 2.94 \\
    \midrule
    \multirow{3}{*}{\textbf{UltraMedical3.1-8B}} 
    & \textbf{GTPA@1} & 0.26 & 0.33 & 0.85 & 0.81 & 0.18 & 0.16 & 0.18 & 0.15 & 0.22 & 0.24 & 0.60& 0.57 \\
    & \textbf{GTPA@5} & 0.45 & 0.51 & 0.89 & 0.93 & 0.25 & 0.26 & 0.26 & 0.24 & 0.42 & 0.36 & 0.73 & 0.63 \\
    & \textbf{Avg Rank} & 6.86 & 6.88 & 3.04 & 2.09 & 8.45 & 8.52 & 8.24 & 8.70& 7.02 & 7.21 & 3.66 & 4.66 \\
    
    \midrule
    \midrule
    \multicolumn{14}{l}{\textit{\textbf{Chain-of-Thought (CoT)}}}\\
    \midrule
    \multirow{3}{*}{\textbf{GPT-4o}} 
    & \textbf{GTPA@1} & 0.71 & 0.72 & 0.97 & 0.96 & 0.68 & 0.64 & 0.60 & 0.64 & 0.47 & 0.57 & 0.82 & 0.81 \\
    & \textbf{GTPA@5} & 0.92 & 0.92 & 1.00 & 1.00 & 0.77 & 0.72 & 0.70 & 0.73 & 0.69 & 0.77 & 0.88 & 0.91 \\
    & \textbf{Avg Rank} & 2.10 & 1.98 & 1.03 & 1.05 & 3.35 & 3.79 & 4.00 & 3.68 & 4.02 & 3.48 & 2.11 & 2.04 \\
    \midrule

    \multirow{3}{*}{\textbf{Llama3.1-70B}} 
    & \textbf{GTPA@1} & 0.45 & 0.58 & 0.89 & 0.91 & 0.48 & 0.44 & 0.45 & 0.45 & 0.49 & 0.50 & 0.71 & 0.75 \\
    & \textbf{GTPA@5} & 0.78 & 0.82 & 0.93 & 0.95 & 0.66 & 0.62 & 0.63 & 0.61 & 0.75 & 0.72 & 0.87 & 0.88 \\
    & \textbf{Avg Rank} & 3.69 & 3.08 & 1.71 & 1.55 & 4.50 & 4.88 & 4.90 & 4.93 & 3.91 & 4.04 & 2.62 & 2.35 \\

    \midrule
    \multirow{3}{*}{\textbf{UltraMedical-70B}}  
    & \textbf{GTPA@1} & 0.47 & 0.47 & 0.96 & 0.93 & 0.26 & 0.33 & 0.34 & 0.34 & 0.39 & 0.35 & 0.69 & 0.32 \\
   & \textbf{GTPA@5} & 0.57 & 0.63 & 1.00& 0.99 & 0.26 & 0.42 & 0.38 & 0.41 & 0.62 & 0.43 & 0.78 & 0.47 \\
    & \textbf{Avg Rank} & 5.46 & 6.70& 1.04 & 1.17 & 8.35 & 6.78 & 7.11 & 6.80& 5.05 & 6.75 & 3.36 & 6.53 \\
    \midrule
    \multirow{3}{*}{\textbf{Llama3.1-8B}}  
    & \textbf{GTPA@1} & 0.45 & 0.51 & 0.97 & 0.95 & 0.27 & 0.34 & 0.3 & 0.29 & 0.24 & 0.36 & 0.65 & 0.64 \\
   & \textbf{GTPA@5} & 0.70& 0.71 & 1.00& 0.99 & 0.40& 0.44 & 0.44 & 0.36 & 0.55 & 0.61 & 0.82 & 0.84 \\
    & \textbf{Avg Rank} & 4.51 & 5.08 & 1.03 & 1.19 & 7.25 & 6.45 & 6.66 & 7.28 & 5.65 & 4.98 & 2.95 & 2.96 \\
    \midrule
    \multirow{3}{*}{\textbf{UltraMedical3.1-8B}} 
    & \textbf{GTPA@1} & 0.22 & 0.22 & 0.74 & 0.81 & 0.13 & 0.13 & 0.14 & 0.18 & 0.09 & 0.19 & 0.39 & 0.50 \\
   & \textbf{GTPA@5} & 0.32 & 0.28 & 0.79 & 0.88 & 0.17 & 0.16 & 0.18 & 0.20& 0.17 & 0.31 & 0.49 & 0.58 \\
    & \textbf{Avg Rank} & 15.30 & 16.25 & 7.24 & 3.97 & 9.33 & 9.41 & 9.26 & 9.02 & 9.36 & 7.99 & 6.23 & 5.26 \\
    
    \bottomrule
    \end{tabular}
    \caption{Diagnosis strategy module performance across 3 datasets with different methods (Standard vs. CoT), aggregated over 100 patients.} 
    \label{tab:performance_diagnosis}
\end{table*}

\newpage
\subsection{Optimizing Knowledge Retrieval Agent vs. Diagnosis Strategy Agent}
\setlength{\tabcolsep}{2.4pt}
\begin{table*}[h]
\centering
\scriptsize
\begin{tabular}{rccccccccc}
\toprule
                               & \multicolumn{3}{c}{\textbf{DDxPlus}} & \multicolumn{3}{c}{\textbf{iCraft-MD}} & \multicolumn{3}{c}{\textbf{RareBench}} \\ \cmidrule(lr){2-4} \cmidrule(lr){5-7} \cmidrule(lr){8-10}
                               & \textbf{GTPA@1 $\uparrow$}          & \textbf{GTPA@5 $\uparrow$}   & \textbf{Avg Rank $\downarrow$} & \textbf{GTPA@1 $\uparrow$}       & \textbf{GTPA@5 $\uparrow$}     & \textbf{Avg Rank $\downarrow$}   & \textbf{GTPA@1 $\uparrow$}        & \textbf{GTPA@5 $\uparrow$}   & \textbf{Avg Rank $\downarrow$}     \\\midrule
                               & \multicolumn{9}{c}{\textbf{GPT-4o}}                                                                 \\\midrule
Retrieval (PubMed)                   & 0.69           &  0.90   &  2.27  & 0.68        &     \textbf{0.79}    & 3.23  & 0.45         &   0.72  &    3.92   \\
Retrieval (Wiki)                   &    0.69        &   0.90 &  2.24  & \textbf{0.69}         &     \textbf{0.79}    & \textbf{3.22}  &   0.45       &  0.74  &  4.00    \\ \cmidrule(lr){2-10}
Zero-shot (Standard)                     &     0.69           &     0.90       &      2.21      &       0.68         &      0.77        &         3.37       &       0.46       &      0.72       &   3.99             \\
Zero-shot (CoT)                    &     0.71          &     0.92       &      2.10      &       0.68         &     0.77         &         3.35       &       0.47       &    0.69         &   4.02              \\ 
Few-shot (Standard, Dyn\_BAII)$\ddag$ &      0.96          &      \textbf{1.00}      &    1.06         &        0.62        &        0.72      &     3.85           &     0.79         &   \textbf{0.91}          &      \textbf{2.03}           \\
Few-shot (CoT, Dyn\_BERT)      &       0.96         &    \textbf{1.00}        &  1.05        &     0.64           &      0.73        &         3.68      &     0.81         &  \textbf{0.91}            &          2.04      \\
Few-shot (CoT, Dyn\_BAII)      &       \textbf{0.97}         &     \textbf{1.00}       &       \textbf{1.03}      &         0.60       &      0.70        &        4.00      &         \textbf{0.82}     &      0.88       &         2.11      \\

\midrule
                               & \multicolumn{9}{c}{\textbf{Llama3.1-70B}}                                                           \\ \midrule
Retrieval (PubMed)                 & 0.56           &  0.79     & 3.42 & 0.44        &    0.63     &  4.72 & 0.38         &   0.75       & 3.96 \\
Retrieval (Wiki)                    &   0.49         & 0.77   &  3.60  &       0.44  &  \textbf{0.66}       &  4.71 &     0.39     & 0.75   &  4.09   \\\cmidrule(lr){2-10}
Zero-shot (Standard)                     &      0.54          &    0.78        &       3.53     &     0.40           &     0.64         &        4.87      &      0.39        &       0.77      &         4.05         \\
Zero-shot (CoT)                      &     0.45          &    0.78      &       3.69     &       \textbf{0.48}         &      \textbf{0.66}        &         \textbf{4.50}       &       0.49       &       0.75      &   3.91              \\
Few-shot (Standard, Dyn\_BAII)$\ddag$ &        0.86        &    \textbf{0.95}        &    1.59         &     0.40           &      0.63        &         5.02       &     0.73         &       0.87      &      2.44           \\
Few-shot (CoT, Dyn\_BERT)      &      \textbf{0.91}          &     \textbf{0.95}       &   \textbf{1.55}         &      0.45          &   0.61           &        4.93     &     \textbf{0.75}         &   \textbf{0.88}          &        \textbf{2.35}      \\
Few-shot (CoT, Dyn\_BAII)      &     0.89           &     0.93       &       1.71     &      0.45          &    0.63          &       4.90      &         0.71     &     0.87        &         2.62     \\ \midrule
                               & \multicolumn{9}{c}{\textbf{Llama3.1-8B}}                                                            \\\midrule
Retrieval (PubMed)                   & 0.42           &  0.67   & 4.50  & 0.29        &   \textbf{0.44}   &   6.93   & 0.35         &   0.59   &    5.33  \\
Retrieval (Wiki)                    &    0.43        &  0.67   &  4.56  &  0.29       &   \textbf{0.44}      & 6.93  &     0.36     &   0.67 &    4.80   \\\cmidrule(lr){2-10}
Zero-shot (Standard)                      &    0.45            &      0.68      &   9.00          &    0.27             &    0.43          &      7.02         &    0.33           &    0.57         &             5.45     \\
Zero-shot (CoT)                     &    0.45            &      0.70      &   4.51          &    0.27             &      0.40        &      7.25         &    0.24           &     0.55        &             5.65     \\
Few-shot (Standard, Dyn\_BAII)$\ddag$  &     \textbf{0.97}           &    \textbf{1.00}        &   \textbf{1.03}          &    0.21            &      0.42        &    6.93            &  \textbf{0.71}            &     0.83        &     \textbf{2.80}            \\
Few-shot (CoT, Dyn\_BERT)     &     0.95           &     0.99       &       1.19      &     0.29           &      0.36        &        7.28     &  0.64            &    \textbf{0.84}         &        2.96      \\
Few-shot (CoT, Dyn\_BAII)       &      \textbf{0.97}          &    \textbf{1.00}        &       \textbf{1.03}     &      \textbf{0.30}          &   \textbf{0.44}           &         \textbf{6.66}     &  0.65            &      0.82       &        2.95      \\

\bottomrule
    \end{tabular}
    \caption{Full comparison of knowledge retrieval agent with diagnosis strategy agent, assuming that there are existing \emph{full} patient profiles. $\ddag$ Only Few-shot (Standard, Dyn\_BAII) results are recorded, since the method is consistently better than Dyn\_BERT.}
    \label{tab:with_patient_profile_full}
\end{table*}

\subsection{MEDDxAgent}

\setlength{\tabcolsep}{3.4pt}
\begin{table*}[th]
    \centering
    \scriptsize
    \begin{tabular}{l l ccc ccc ccc ccc ccc ccc}
    \toprule
    & & \multicolumn{6}{c}{\textbf{DDxPlus}} & \multicolumn{6}{c}{\textbf{iCraft-MD}} & \multicolumn{6}{c}{\textbf{RareBench}} \\
    \cmidrule(lr){3-8} \cmidrule(lr){9-14} \cmidrule(lr){15-20}
    & & \multicolumn{3}{c}{\textbf{Fixed}} & \multicolumn{3}{c}{\textbf{Dynamic}} 
      & \multicolumn{3}{c}{\textbf{Fixed}} & \multicolumn{3}{c}{\textbf{Dynamic}} 
      & \multicolumn{3}{c}{\textbf{Fixed}} & \multicolumn{3}{c}{\textbf{Dynamic}} \\
    \cmidrule(lr){3-5} \cmidrule(lr){6-8} \cmidrule(lr){9-11} \cmidrule(lr){12-14} \cmidrule(lr){15-17} \cmidrule(lr){18-20}
    \textbf{Model} & \textbf{Metric} & \textbf{1} & \textbf{2} & \textbf{3} & \textbf{1} & \textbf{2} & \textbf{3} 
                                       & \textbf{1} & \textbf{2} & \textbf{3} & \textbf{1} & \textbf{2} & \textbf{3} 
                                       & \textbf{1} & \textbf{2} & \textbf{3} & \textbf{1} & \textbf{2} & \textbf{3} \\
    \midrule

    \multirow{3}{*}{\textbf{GPT-4o}} 
    & \textbf{GTPA@1}  & 0.74 & 0.78 & 0.86 & 0.74 & 0.76 & 0.81 & 0.52 & 0.54 & 0.54 & 0.44 & 0.52 & 0.52 & 0.51 & 0.56 & 0.50 & 0.35 & 0.43 & 0.46 \\
    & \textbf{Avg Rank} & 1.91 & 1.56 & 1.29 & 2.00 & 1.62 & 1.48 & 4.93 & 4.71 & 4.80 & 4.85 & 4.57 & 4.56 & 4.37 & 4.10 & 4.09 & 6.14 & 4.62 & 4.24 \\
    & \textbf{Avg Progress} & 0.00 & 0.32 & 0.32 & -0.13 & 0.04 & 0.01 & 0.00 & 0.26 & 0.17 & 0.14 & -0.06 & 0.01 & 0.00 & 0.13 & 0.16 & -0.23 & -0.15 & -0.39 \\
    \midrule

    \multirow{3}{*}{\textbf{Llama3.1-70B}} 
    & \textbf{GTPA@1}  & 0.61 & 0.71 & 0.68 & 0.53 & 0.61 & 0.60 & 0.29 & 0.37 & 0.42 & 0.24 & 0.30 & 0.31 & 0.39 & 0.48 & 0.48 & 0.32 & 0.37 & 0.46 \\
    & \textbf{Avg Rank} & 2.91 & 2.20 & 2.30 & 2.96 & 2.89 & 2.73 & 7.05 & 6.26 & 6.31 & 7.08 & 6.77 & 6.82 & 5.05 & 4.48 & 4.30 & 5.44 & 4.66 & 4.19 \\
    & \textbf{Avg Progress} & 0.00& 0.41 & 0.17 & 0.00& 0.04 & 0.02 & 0.00& 0.07 & 0.26 & 0.10 & 0.02 & 0.00& 0.00& 0.75 & 0.44 & -0.06 & 0.00& 0.07 \\
    \midrule
    \multirow{3}{*}{\textbf{Llama3.1-8B}} 
    & \textbf{GTPA@1}  & 0.34 & 0.56 & 0.58 & 0.47 & 0.58 & 0.54 & 0.11 & 0.14 & 0.12 & 0.03 & 0.04 & 0.04 & 0.08 & 0.09 & 0.07 & 0.06 & 0.10 & 0.18   \\
    & \textbf{Avg Rank} & 5.25 & 3.59 & 3.10 & 5.00 & 3.82 & 3.92 & 9.38 & 9.22 & 9.07 & 10.11 & 9.95 & 9.91 & 8.47 & 8.11 & 8.56 & 8.01 & 7.54 & 7.21 \\
    & \textbf{Avg Progress} & 0.00& 1.73 & 1.23 & 0.00& 0.00& 0.00& 0.00& 0.22 & 0.17 & 0.00& 0.00& 0.00& 0.00& 0.44 & 0.38 & 0.00& 0.00& 0.00\\

\bottomrule
    \end{tabular}
    \caption{Iterative experiment performance compared between fixed iteration and dynamic iteration with 3 datasets.}
    \label{tab:choice_merged}
\end{table*}

\end{document}